\documentclass{article}

\usepackage[resetlabels,labeled]{multibib}
\newcites{S}{Appendix References} 

\usepackage[final]{neurips_2022}

\usepackage[utf8]{inputenc} 
\usepackage[T1]{fontenc}    
\usepackage{url}            
\usepackage{booktabs}       
\usepackage{amsfonts}       
\usepackage{nicefrac}       
\usepackage{microtype}      
\usepackage{xspace}
\usepackage{url}
\usepackage{graphicx}
\usepackage{bm}
\usepackage{amssymb}
\usepackage{amsmath}
\usepackage{amsthm}
\usepackage{times}
\usepackage[ruled,vlined]{algorithm2e}
\usepackage[table,xcdraw]{xcolor}
\usepackage[title]{appendix}
\usepackage{wrapfig}
\usepackage[colorlinks=true,linkcolor=gray, urlcolor=black,citecolor=gray, anchorcolor=blue]{hyperref}
\usepackage{comment}
\usepackage{bbm}
\usepackage{multirow}

\newcommand{\ziyuan}[1]{{\color{cyan}[Z:{#1}]}}

\newcommand{\revise}[1]{{\color{black}#1}}

\newcommand{\ie}{\textit{i.e.}}
\newcommand{\eg}{\textit{e.g.}}
\newcommand{\etal}{\textit{et al.}\xspace}
\newcommand{\cut}[1]{}  
\newcommand{\std}[1]{\scriptsize{$\pm$#1}}
\newcommand{\xhdr}[1]{\vspace{-1mm}\noindent{{\bf #1.}}}
\newcommand{\xhdrnm}[1]{\noindent{{\bf #1.}}}

\newtheorem*{problem*}{Problem (Self-Supervised Contrastive Pre-Training For Time Series)}

\newtheorem*{TFC*}{Representational Time-Frequency Consistency (TF-C)}

\usepackage[font=small,labelfont=bf]{caption}
\newcommand{\Hquad}{\hspace{0.3em}} 
 
\usepackage{array}
\newcolumntype{H}{>{\setbox0=\hbox\bgroup}c<{\egroup}@{}}

\usepackage{contour}
\usepackage{ulem}

\contourlength{0.8pt}
\newcommand{\myuline}[1]{%
  \uline{\phantom{#1}}%
  \llap{\contour{white}{#1}}%
}
\renewcommand{\underline}{\myuline}

\title{Self-Supervised Contrastive Pre-Training for Time Series via Time-Frequency Consistency}

\author{
Xiang Zhang$^{\dagger}$ $^*$\\
Harvard University\\
\texttt{xiang\_zhang@hms.harvard.edu} \\
\And
Ziyuan Zhao$^*$\\
Harvard University\\
\texttt{ziyuanzhao@college.harvard.edu} \\
\And
\: \: \: \:  Theodoros Tsiligkaridis\\
\: \: \: \: MIT Lincoln Laboratory\\
\: \: \: \: \texttt{ttsili@ll.mit.edu} \\
\And
\: \: \: \: \:  Marinka Zitnik\\
\: \: \: \: \: Harvard University\\
\: \: \: \: \: \texttt{marinka@hms.harvard.edu} \\
}

\begin{document}

\maketitle
\def\thefootnote{*}\footnotetext{These authors contributed equally to this work.
} \def\thefootnote{\arabic{footnote}}
\def\thefootnote{$\dagger$}\footnotetext{Present address: University of North Carolina at Charlotte, xiang.zhang@uncc.edu}
\def\thefootnote{\arabic{footnote}}
\begin{abstract}
Pre-training on time series poses a unique challenge due to the potential mismatch between pre-training and target domains, such as shifts in temporal dynamics, fast-evolving trends, and long-range and short-cyclic effects, which can lead to poor downstream performance. While domain adaptation methods can mitigate these shifts, most methods need examples directly from the target domain, making them suboptimal for pre-training. 
To address this challenge, methods need to accommodate target domains with different temporal dynamics and be capable of doing so without seeing any target examples during pre-training. Relative to other modalities, in time series, we expect that time-based and frequency-based representations of the same example are located close together in the time-frequency space. To this end, we posit that time-frequency consistency (TF-C) --- embedding a time-based neighborhood of an example close to its frequency-based neighborhood --- is desirable for pre-training.
Motivated by TF-C, we define a decomposable pre-training model, where the self-supervised signal is provided by the distance between time and frequency components, each individually trained by contrastive estimation. 
We evaluate the new method on eight datasets, including electrodiagnostic testing, human activity recognition, mechanical fault detection, and physical status monitoring.  
Experiments against eight state-of-the-art methods show that TF-C outperforms baselines by 15.4\% (F1 score) on average in one-to-one settings (\eg, fine-tuning an EEG-pretrained model on EMG data) and by 8.4\% (precision) in challenging one-to-many settings (\eg, fine-tuning an EEG-pretrained model for either hand-gesture recognition or mechanical fault prediction), reflecting the breadth of scenarios that arise in real-world applications.
The source code and datasets are available at \url{https://github.com/mims-harvard/TFC-pretraining}.

\end{abstract}

\section{Introduction}\label{sec:intro}

Time series plays important roles in many areas, including clinical diagnosis, traffic analysis, and climate science~\cite{harutyunyan_multitask_2019, rezaei_deep_2019, ravuri_skilful_2021, sezer_financial_2020,su2021temporal,deng2021deep}. While representation learning has considerably advanced analysis of time series~\cite{rebjock2021online,sun2021adjusting,dempster2020rocket} more broadly~\cite{huang2022spiral}, learning generalizable representations for temporal data remains a fundamentally challenging problem~\cite{sun2021adjusting,ismail2019deep}. There are numerous immediate benefits from generating such representations, of which pre-training capability is particularly desirable and of great practical importance~\cite{shi_self-supervised_2021,dang2021ts}. Central to pre-training is a question of how to process time series in a diverse dataset to greatly improve generalization on new time series coming from different datasets~\cite{changpinyo2021conceptual,sun2021multilingual,huang2022spiral}. By training a neural network model on a
dataset and transferring it to a new target dataset for fine-tuning, \ie, without explicit retraining on that target data, we expect the resulting performance to be at least as good as that of state-of-the-art models tailored to the target dataset. 

However, unfortunately, the expected performance gains are often not realized for a variety of reasons (\eg, distribution shifts, properties of the target dataset unknown during pre-training)~\cite{ye2021implementing,fawaz2018transfer} that get compounded by the complexity of time series: large variations of temporal dynamics across datasets, varying semantic meaning, irregular sampling, system factors (\eg, different devices or subjects), etc.~\cite{wickstrom_mixing_2022,fawaz2018transfer}.   
This complexity of time series limits the utility of knowledge transfer for pre-training~\cite{gupta2020transfer,meiseles2020source}. For example, pre-training a model on a diverse time series dataset with mostly low-frequency components (smooth trends) may not lead to positive transfer on downstream tasks with high-frequency components (transient events)~\cite{fawaz2018transfer}. Examining these challenges can provide clues to what kind of inductive biases could facilitate generalizable representations of time series -- this paper offers a strategy for that through a novel time-frequency consistency principle.  

\begin{figure}
    \centering
    \includegraphics[width=\linewidth]{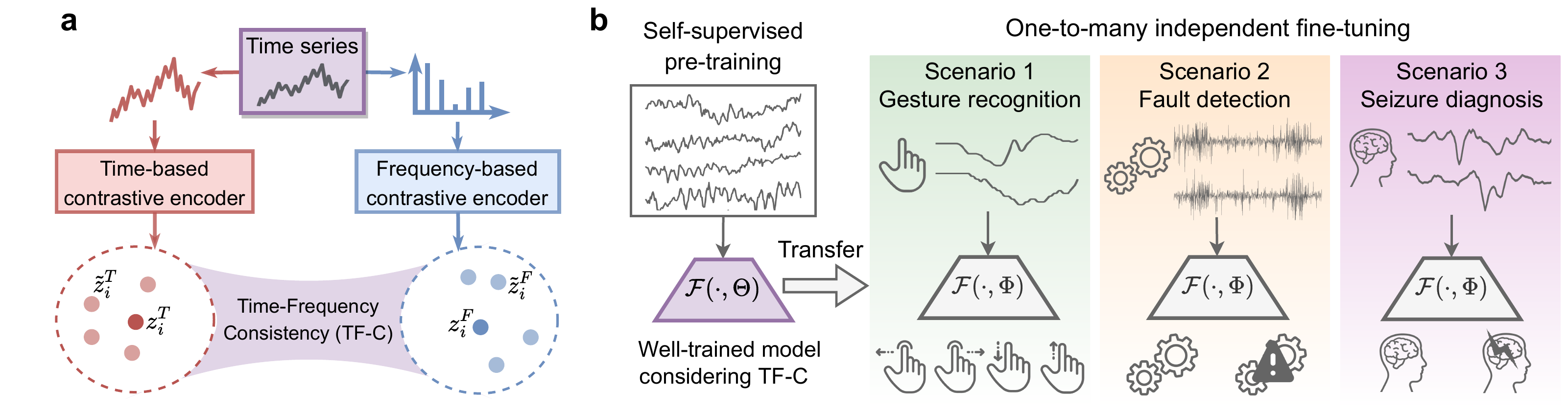}
    \caption{
    \textbf{a.} Illustration of Time-Frequency Consistency (TF-C). Time-based embedding $\bm{z}^{\textsc{t}}_i$ and frequency-based embedding $\bm{z}^{\textsc{f}}_i$ of time series sample $\bm{x}^{\textsc{t}}_i$, along with $\widetilde{\bm{z}}^{\textsc{t}}_i$ and $\widetilde{\bm{z}}^{\textsc{f}}_i$ learned from augmentations of $\bm{x}^{\textsc{t}}_i$, should be close to each other in the latent time-frequency space. 
    \textbf{b.} Leveraging TF-C property in time series to optimize a pre-training model $\mathcal{F}$ with parameters $\Theta$ that get fine-tuned to $\Phi$ on a small scenario-specific dataset.
    }
    \label{fig:TS_pretrain}
\vspace{-8mm}
\end{figure}

In addition, target datasets are not available during pre-training (different from domain adaption~\cite{singh2021clda}; Appendix~\ref{app:difference_domain_adaptation}), requiring that the pre-training model captures a latent property that holds true for previously unseen target datasets. 
At the center of this desideratum is the idea of a property that would be shared between pre-training and target datasets and would enable knowledge transfer from pre-training to fine-tuning. In computer vision (CV), pre-training is driven by findings that initial neural layers capture universal visual elements, such as edges and shapes, that are relevant regardless of image style and tasks~\cite{geirhos_imagenet-trained_2018}. In natural language processing (NLP), the foundation for pre-training is given by linguistic principles of semantics and grammar shared across different languages~\cite{Radford2018ImprovingLU}.
However, due to the aforementioned temporal complexity, such a principle for pre-training on time series has not yet been established.
Moreover, supervised pre-training requires access to large annotated datasets, which limits its use in domains where richly labeled datasets are scarce~\cite{Chen:neurips:2020,wav2vec:2020}. 
For example, in medical applications, labeling data at scale is often infeasible or can be expensive and noisy (experts can disagree on ground-truth labeling~\cite{clifford_af_2017,gordon2021disagreement}, \eg, whether an ECG signal indicates a normal vs.~abnormal rhythm)~\cite{rogers2013investigating,horowitz1981prototype}.
To mitigate these issues, self-supervised learning emerged as a promising strategy to sidestep the lack of labeled datasets~\cite{oord_representation_2018}.

\xhdr{Present work}
We introduce a strategy for self-supervised pre-training in time series by modeling Time-Frequency Consistency (TF-C). TF-C specifies that time-based and frequency-based representations, learned from the same time series sample, should be closer to each other in the time-frequency space than representations of different time series samples.
Specifically, we adopt contrastive learning in time-space to generate a time-based representation. In parallel, we propose a set of novel augmentations based on the characteristic of the frequency spectrum and produce a frequency-based embedding through contrastive instance discrimination. This is the first work to develop frequency-based contrastive augmentation to leverage rich spectral information and explore time-frequency consistency in time series.
%
The pre-training objective is to minimize the distance between time-based and frequency-based embeddings using a novel consistency loss (Figure~\ref{fig:TS_pretrain} (a)). The self-supervised loss is used to optimize the pre-training model and enforce consistency between time and frequency domains in the latent space. 
The learned relationship encoded in model parameters are transferred to initialize the fine-tuning model and improve performance in datasets of interest (Figure~\ref{fig:TS_pretrain} (b)).

We evaluate the TF-C model on eight time series datasets under two evaluation settings (\ie, one-to-one and one-to-many). The eight datasets cover a large set of variations: different numbers of channels (from univariate to 9-channel multivariate), varying time series lengths (from 128 to 5,120), different sampling rates (from 16 Hz to 4,000 Hz), different scenarios (neurological healthcare, human activity recognition, mechanical fault detection, physical status monitoring, etc.) and diverse types of signals (EEG, EMG, ECG, acceleration, and vibration). 
We compare TF-C approach to eight state-of-the-art baselines. Results show that TF-C achieves positive transfer, outperforming all baselines by a large margin of 15.4\% (F1 score) on average.
Further, the approach outperforms the strongest baselines with an improvement of up to 7.2\% in the F1 score. 
Finally, the TF-C approach improves prior work by 8.4\% in precision (when pre-training the model on sleep EEG signals and fine-tuning it on hand-gesture recognition) in challenging one-to-many setups that apply the same pre-trained model to multiple independent fine-tuning datasets. 

\vspace{-3mm}

\section{Related Work}\label{sec:related}

\xhdr{Pre-training for time series} 
Although there are studies on self-supervised representation learning for time series~\cite{rebjock2021online,sun2021adjusting,Sarkar:2020,cheng_subject-aware_2020} and self-supervised pre-training for images~\cite{ravula2021inverse,chen_simple_2020,dai_up-detr_2021,Chen:neurips:2020}, the intersection of these two areas, \ie, self-supervised pre-training for time series, remains underexplored. In time series, it's not obvious what reasonable assumptions can bridge pre-training and target datasets. Hence, pre-training models in CV~\cite{lee_unsupervised_2017,caron2019unsupervised,changpinyo2021conceptual} and NLP~\cite{huang2022spiral,sun2021multilingual,devlin_bert_2019} are not directly applicable due to data modality mismatch, and the existing results leave room for improvement~\cite{Sarkar:2020,Wu:2020,tang_exploring_2021}.
Shi \etal~\cite{shi_self-supervised_2021} developed the only model to date that is explicitly designed for self-supervised time series pre-training. The model captures the local and global temporal pattern, but it is not convincing why the designed pretext task can capture generalizable representations.
Although several studies applied transfer learning in the context of time series~\cite{rebjock2021online,sun2021adjusting,wickstrom_mixing_2022,kiyasseh_clocs_nodate}, there is no foundation yet of which conceptual properties are most suitable for pre-training on time series and why. Addressing this gap, we show that TF-C, designed to be invariant to different time-series datasets, can produce generalizable pre-training models.


Unlike domain adaptation~\cite{singh2021clda,berthelot2021adamatch} that requires access to target datasets during training, pre-training models do not have access to fine-tuning datasets. As a result, one needs to identify a generalizable time-series property to benefit from pre-training. Further, self-supervised domain adaptation does not need labels in the target dataset but still requires labels for model training~\cite{wei2021toalign,xu2021cdtrans}. In contrast, TF-C does not need any labels during pre-training.

\xhdrnm{Contrastive learning with time series}
%
Contrastive learning, a popular type of self-supervised learning, aims to learn an encoder that maps inputs into an embedding space such that positive sample pairs (original augmentation and another alternative augmentation/view of the same input sample) are pulled closer and negative sample pairs (original augmentation and an alternative input sample augmentation) are pushed apart~\cite{oord_representation_2018,illing2021local}.
Contrastive learning in time series is less investigated in comparison, partly due to the challenge of identifying augmentations that capture
key invariance properties in time series data. For example, CLOCS defines adjacent time segments as positive pairs \cite{kiyasseh_clocs_nodate}, and TNC assumes overlapping temporal neighborhoods have similar representations \cite{tonekaboni_unsupervised_2021}. 
These methods leverage temporal invariance to define positive pairs which are used to calculate contrastive loss, but other invariances, such as transformation invariance (\eg, SimCLR~\cite{tang_exploring_2021}), contextual invariance (\eg, TS2vec \cite{yue_ts2vec_2022} and TS-TCC \cite{eldele_time-series_2021}) and augmentations are possible. 
In this work, we propose an augmentation bank that exploits multiple invariances to generate diverse augmentations (Sec.~\ref{sub:time_contrastive}), which adds richness to the pre-training model~\cite{eldele_time-series_2021}.
Importantly, we propose frequency-based augmentations by perturbing the frequency spectrum of time series (\eg, adding or removing the frequency components and manipulating their amplitude; more details in Sec.~\ref{sub:frequency_contrastive}) to learn better representations by exposing the model to a local range of frequency variations. In previous work, CoST processes sequential signals through the frequency domain, but the augmentations are still implemented in time space~\cite{woo_cost_2022}. 
\revise{Similarly, although BTSF~\cite{yang2022unsupervised} involves frequency domain, its data transformation is solely implemented in the time domain using instance-level dropout.} Additional commentary on differences between CoST and BTSF is in Appendix~\ref{app:BTSF}.
To the best of our knowledge, \revise{this is the first work that directly perturbs the frequency spectrum to leverage frequency-invariance for contrastive learning.}
Further, we develop a pre-training model that subjects to TF-C upon two individual contrastive encoders.

\section{Problem Formulation}\label{sec:problem_formulation}

We are given a pre-training dataset $\mathcal{D}^{\textrm{pret}} = \{ \bm{x}^{\textrm{pret}}_i \;|\; i=1, \dots, N \}$ of unlabeled time series samples where sample $\bm{x}^{\textrm{pret}}_i$ has $K^{\textrm{pret}}$ channels and $L^{\textrm{pret}}$ timestamps.
Let $\mathcal{D}^{\textrm{tune}} = \{(\bm{x}^{\textrm{tune}}_i, y_i) \;|\; i=1, \dots, M\}$ be a fine-tuning (\ie, target; target and fine-tuning are used interchangeably) dataset of labeled time series samples, each having  $K^{\textrm{tune}}$ channels and $L^{\textrm{tune}}$ timestamps. Furthermore, every sample $\bm{x}^{\textrm{tune}}_i$ is associated with a label $y_i \in \{1, \dots, C\}$, where $C$ is the number of classes. 
Without loss of generality, in the following descriptions, we focus on univariate (single-channel) time series, while noting that our approach can accommodate multivariate time series of varying lengths across datasets
(shown in experiments in Sec.~\ref{sub:one_to_many}).  
We use superscript symbol $\Hquad \widetilde{} \Hquad$ to denote contrastive augmentations. We note that $\bm{x}^{\textsc{t}}_i \equiv \bm{x}_i$ denotes an input time series sample, and $\bm{x}^{\textsc{f}}_i$ denotes discrete frequency spectrum of $\bm{x}_i$.
%

\begin{problem*} 
Given are an unlabeled pre-training dataset $\mathcal{D}^{\textrm{pret}}$ with $N$ samples and a target dataset $\mathcal{D}^{\textrm{tune}}$ with $M$ samples ($M \ll N$).
The goal is to use $\mathcal{D}^{\textrm{pret}}$ to pre-train a model $\mathcal{F}$ so that by fine-tuning model parameters on $\mathcal{D}^{\textrm{tune}}$, the fine-tuned model produces generalizable representations $\bm{z}^{\textrm{tune}}_i = \mathcal{F} (\bm{x}^{\textrm{tune}}_i)$ for every $\bm{x}^{\textrm{tune}}_i$. 
\end{problem*} 

We follow an established setup, \eg,~\cite{kiyasseh_clocs_nodate}: for pre-training, only the unlabeled dataset $\mathcal{D}^{\textrm{pret}}$ is available while, for  fine-tuning, a small labeled dataset $\mathcal{D}^{\textrm{tune}}$ can be used. In short, a model $\mathcal{F}$ is pre-trained on the unlabeled time series dataset $\mathcal{D}^{\textrm{pret}}$ and its optimized model parameters $\Theta$ are fine-tuned to go from $\mathcal{F}(\cdot, \Theta)$ to $\mathcal{F}(\cdot, \Phi)$ using the dataset $\mathcal{D}^{\textrm{tune}}$. The $\Phi$ denotes fine-tuned model parameters.
Note that this problem (\ie, $\mathcal{D}^{\textrm{pret}}$ is independent of the target dataset) is distinct from domain adaptation as fine-tuning dataset $\mathcal{D}^{\textrm{tune}}$ is not accessed during pre-training. As a result, the pre-trained model can be used with many different fine-tuning datasets without re-training.

\xhdr{Rationale for Time-Frequency Consistency (TF-C)} 
The central idea is to identify a general property that is preserved across time series datasets and use it to induce transfer learning for effective pre-training. The time domain shows how sensor readouts change with time, whereas the frequency domain shows how much of the signal lies within each frequency component over the entire spectrum~\cite{hyndman_forecasting_2018}. Explicitly considering the frequency domain can provide an understanding of time series behavior that cannot be directly captured solely in the time domain~\cite{bracewell1986fourier}. However, existing contrastive methods (\eg,~\cite{yue_ts2vec_2022,eldele_time-series_2021}) focus exclusively on modeling the time domain and ignore the frequency domain altogether. One can argue that approach is sufficient in the case of high-capacity methods as time and frequency domains are different views of the same data~\cite{cohen_time-frequency_1995}, which can be cross-translated using transformation, such as Fourier and inverse Fourier~\cite{nussbaumer1981fast,bracewell1986fourier}. 
The relationship between the two domains, grounded in signal processing theory, provides an invariance that is valid regardless of the time series distribution~\cite{flandrin_time-frequencytime-scale_1998, papandreou-suppappola_applications_2018} and thus can serve as an inductive bias for pre-training. 
Appendix~\ref{app:invariance} provides a commentary with analogies for images. 
Approaching this invariance through the lens of representation learning, we next formulate Time-Frequency Consistency (TF-C). The TF-C property postulates there exists a latent time-frequency space such that for every sample $\bm{x}_i$, time-based representation $\bm{z}^{\textsc{t}}_i$ and frequency-based representation $\bm{z}^{\textsc{f}}_i$ of the same sample, together with their local augmentations (defined later), are close to each other in the latent space. 
%
\begin{TFC*}
Let $\bm{x}_i$ be a time series and $\mathcal{F}$ be a model satisfying TF-C. Then, time-based representation $\bm{z}^\textsc{t}_i$ and frequency-based representation $\bm{z}^\textsc{f}_i$ as well as representations of $\bm{x}_i$'s local augmentations are proximal in the latent time-frequency space.
\end{TFC*}
Our strategy is to use dataset $\mathcal{D}^{\textrm{pret}}$ to induce TF-C in $\mathcal{F}$'s model parameters $\Theta$, which, in turn, are used to initialize the target model on \revise{$\mathcal{D}^{\textrm{tune}}$} and produce generalizable representations for downstream prediction. 
The invariant nature of TF-C means that the approach can bridge $\mathcal{D}^{\textrm{pret}}$ and  \revise{$\mathcal{D}^{\textrm{tune}}$} even when large discrepancies exist between them (in terms of temporal dynamics, semantic meaning, etc.), providing a vehicle for a general pre-training on time series. 
%
%

\begin{figure}
    \centering
    \includegraphics[width=\textwidth]{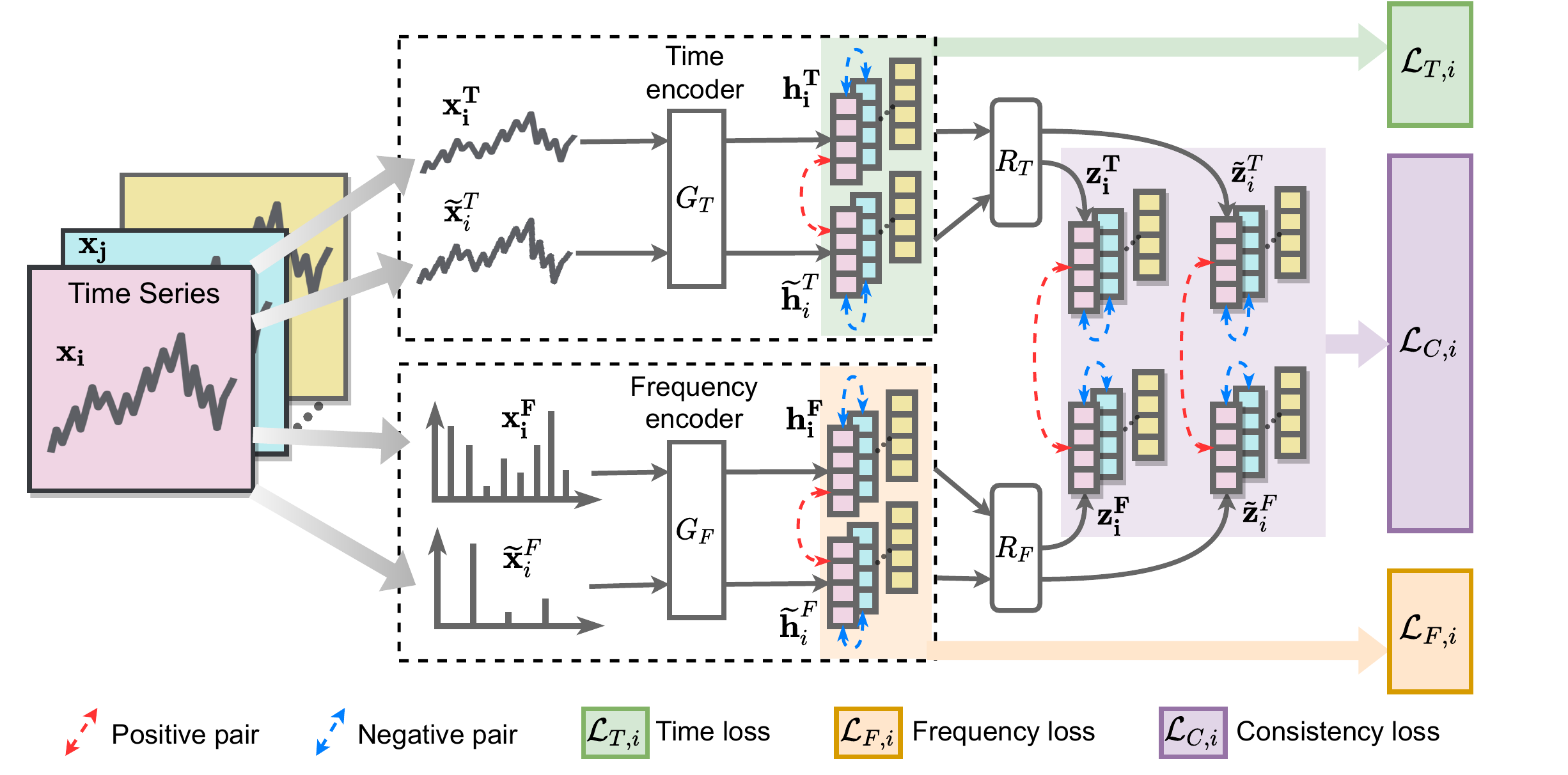}
    \caption{\textbf{Overview of TF-C approach.} Our TF-C pre-training model $\mathcal{F}$ has four components: a time encoder $G_{\textsc{t}}$, a frequency encoder $G_{\textsc{f}}$, and two cross-space projectors $R_{\textsc{t}}$ and $R_{\textsc{f}}$. 
    For an input time series $\bm{x}_i$, the model produces time-based representations (\ie, $\bm{z}^{\textsc{t}}_i$ and $\widetilde{\bm{z}}^{\textsc{t}}_i$ of input $\bm{x}_i$ and its augmented version, respectively) and frequency-based representations (\ie, $\bm{z}^{\textsc{f}}_i$ and $\widetilde{\bm{z}}^{\textsc{f}}_i$ of input $\bm{x}_i$ and its augmented version, respectively). The TF-C property is realized by promoting the alignment of time- and frequency-based representations in the latent time-frequency space, providing a vehicle for transferring $\mathcal{F}$ to a target dataset not seen before.
    }
    \label{fig:framework}
    \vspace{-5mm}
\end{figure}

To realize TF-C, our model $\mathcal{F}$ has four components (Figure~\ref{fig:framework}): a time encoder $G_{\textsc{t}}$, a frequency encoder $G_{\textsc{f}}$, and two cross-space projectors $R_{\textsc{t}}$ and $R_{\textsc{f}}$ that map time-based and frequency-based representations, respectively, to the same time-frequency space.
Together, the four components provide a way to embed $\bm{x}_i$ to the latent time-frequency space such that the time-based embedding $\bm{z}^{\textsc{t}}_i = R_{\textsc{t}}(G_{\textsc{t}}(\bm{x}^{\textsc{t}}_i))$ and the frequency-based embedding  $\bm{z}^{\textsc{f}}_i= R_{\textsc{f}}(G_{\textsc{f}}(\bm{x}^{\textsc{f}}_i))$ are close together. 

\cut{
}

\section{Our Approach}
\label{sec:method}

Next, we present the architecture of the developed self-supervised contrastive pre-training model $\mathcal{F}$.
Unless specified otherwise, the data mentioned in this section are from pre-training dataset and the superscript $^{\textrm{pret}}$ is omitted for simplification. Here we describe the model using univariate time series as an example, but our model can be straightforwardly applied to multivariate time series (Sec~\ref{sec:experiments}).  

\subsection{Time-based Contrastive Encoder}
\label{sub:time_contrastive}

%
For a given input time series sample $\bm{x}_i$, we generate an augmentation set $\mathcal{X}^{\textsc{t}}_i$ through a time-based augmentation bank $\mathcal{B}^{\textsc{t}}: \bm{x}^{\textsc{t}}_i \rightarrow \mathcal{X}^{\textsc{t}}_i$. Each element $\widetilde{\bm{x}}^{\textsc{t}}_i \in  \mathcal{X}^{\textsc{t}}_i$ is augmented from $\bm{x}_i$ based on the temporal characteristics. 
Here, the time-based augmentation bank includes jittering, scaling, time-shifts, and neighborhood segments, all well-established in contrastive learning~\cite{tang_exploring_2021, eldele_time-series_2021,kiyasseh_clocs_nodate}. We develop an augmentation bank to produce diverse augmentations (rather than a single type of augmentation) and expose the model to complex temporal dynamics, which produces more robust time-based embeddings~\cite{eldele_time-series_2021}.

For the input $\bm{x}_i$, we randomly select an augmented sample $\widetilde{\bm{x}}^{\textsc{t}}_i \in  \mathcal{X}^{\textsc{t}}_i$ and feed into a contrastive time encoder $G_{\textsc{t}}$ that maps samples to embeddings. We have  $\bm{h}^{\textsc{t}}_i = G_{\textsc{t}}(\bm{x}^{\textsc{t}}_i)$ and $\widetilde{\bm{h}}^{\textsc{t}}_i = G_{\textsc{t}}(\widetilde{\bm{x}}^{\textsc{t}}_i)$. 
%
As $\widetilde{\bm{x}}^{\textsc{t}}_i$ is generated based on $\bm{x}^{\textsc{t}}_i$, after passing through $G_{\textsc{t}}$, we assume the embedding of $\bm{x}^{\textsc{t}}_i$ is close to the embedding of $\widetilde{\bm{x}}^{\textsc{t}}_i$ but far away from the embedding of $\bm{x}^{\textsc{t}}_j$ and $\widetilde{\bm{x}}^{\textsc{t}}_j$ that are derived from another sample $\bm{x}^{\textsc{t}}_j \in \mathcal{D}^{\textrm{pret}}$~\cite{chen_simple_2020,yue_ts2vec_2022,kiyasseh_clocs_nodate}. In specific, we select the positive pair as $(\bm{x}^{\textsc{t}}_i, \widetilde{\bm{x}}^{\textsc{t}}_i)$ and negative pairs as $(\bm{x}^{\textsc{t}}_i, \bm{x}^{\textsc{t}}_j)$ and $(\bm{x}^{\textsc{t}}_i, \widetilde{\bm{x}}^{\textsc{t}}_j)$~\cite{chen_simple_2020}. 

\xhdr{Contrastive time loss} To maximize the similarity within a positive pair and minimize the similarity within a negative pair, we adopt the NT-Xent (the normalized temperature-scaled cross entropy loss) as distance function $d$ which is widely used in contrastive learning~\cite{chen_simple_2020,tang_exploring_2021}. In specific, we define the loss function of the time-based contrastive encoder in terms of sample $\bm{x}^{\textsc{t}}_i$ as:
\begin{equation}
\label{eq:time_loss}
\mathcal{L}_{\textsc{t}, i} = d(\bm{h}^{\textsc{t}}_i, \widetilde{\bm{h}}^{\textsc{t}}_i, \mathcal{D}^{\textrm{pret}}) = -\textrm{log}\frac{\textrm{exp}(\textrm{sim}(\bm{h}^{\textsc{t}}_i, \widetilde{\bm{h}}^{\textsc{t}}_i)/\tau )}{\sum^{}_{\bm{x}_j \in \mathcal{D}^{\textrm{pret}}} \mathbbm{1}_{i \neq j}\textrm{exp}(\textrm{sim}(\bm{h}^{\textsc{t}}_i,G_{\textsc{t}}(\bm{x}_j))/\tau )},
\end{equation}
where  $\textrm{sim} (\bm{u}, \bm{v})= \bm{u}^T \bm{v}/ \left \|\bm{u}  \right \| \left \|\bm{v}  \right \|$ denotes the cosine similarity, the $\mathbbm{1}_{i \neq j} $ is an indicator function that equals to 0 when $i = j$ and 1 otherwise, and $\tau$ is a temporal parameter to adjust scale. 
The $\bm{x}_j \in \mathcal{D}^{\textrm{pret}}$ refers to a different time series sample or its augmented sample.
This loss function urges the time encoder $G_{\textsc{t}}$ to generate closer time-based embeddings for positive pairs and push the embeddings for negative pairs apart from each other.

\subsection{Frequency-based Contrastive Encoder}
\label{sub:frequency_contrastive}
We generate the frequency spectrum $\bm{x}^{\textsc{f}}_i$ from a time series sample $\bm{x}^\textsc{t}_i$
through a transform operator (\eg, Fourier Transformation~\cite{nussbaumer1981fast}). The frequency information in time series is universal and plays a key role in classic signal processing~\cite{soklaski2022fourier,cohen_time-frequency_1995,flandrin_time-frequencytime-scale_1998}, but it is rarely investigated in self-supervised contrastive representation learning for time series~\cite{jaiswal_survey_2020}. In this section, we develop augmentation method to perturb $\bm{x}^{\textsc{f}}_i$ based on characteristics of frequency spectra and show how to generate frequency-based representations. 

As every frequency component in the frequency spectrum denotes a basis function (\eg, sinusoidal function for Fourier transformation) with the corresponding frequency and amplitude, we perturb the frequency spectrum by adding or removing frequency components.
A small perturbation in the frequency domain may cause large changes to the temporal patterns in the time domain~\cite{flandrin_time-frequencytime-scale_1998}. To make sure the perturbed time series is still similar to the original sample (not only in frequency domain but also in time domain; \revise{Figure~\ref{fig:freq_aug}}), we use a small budget $E$ in the perturbations where $E$ denotes the number of frequency components we manipulate. While removing frequency components, we randomly select $E$ frequency components and set their amplitudes to $0$. While adding frequency components, we randomly choose $E$ frequency components from the ones have smaller amplitude than $\alpha \cdot A_m$, and increase their amplitude to $\alpha \cdot A_m$. The $A_m$ is the maximum amplitude in the frequency spectrum and $\alpha$ is a pre-defined coefficient to adjust the \revise{scale} of the perturbed frequency component ($\alpha=0.5$ in this work).
We produce an augmentation set $\mathcal{X}^\textsc{f}_i$ for $\bm{x}^\textsc{f}_i$ through frequency-augmentation bank $\mathcal{B}^\textsc{f}: \bm{x}^\textsc{f}_i \rightarrow \mathcal{X}^\textsc{f}_i$. As described above, we have two augmentation methods (\ie, removing or adding frequency components) in $\mathcal{B}^\textsc{f}$, $|\mathcal{X}^\textsc{f}_i|= 2$. Details on the exploration of frequency augmentation strategies are covered in Appendix~\ref{app:domain_augmentation}.

We utilize a frequency encoder $G_{\textsc{f}}$ to map the frequency spectrum (\eg, $\bm{x}^{\textsc{f}}_i$) to a frequency-based embedding (\eg, $\bm{h}^{\textsc{f}}_i = G_{\textsc{f}}(\bm{x}^{\textsc{f}}_i)$). We assume the frequency encoder $G_{\textsc{f}}$ can learn similar embedding for the original frequency spectrum $\bm{x}^{\textsc{f}}_i$ and a slightly perturbed frequency spectrum $\widetilde{\bm{x}}^{\textsc{f}}_i \in \mathcal{X}^{\textsc{f}}_i$. Thus, we set the positive pair as $(\bm{x}^{\textsc{f}}_i, \widetilde{\bm{x}}^{\textsc{f}}_i)$ and the negative pairs as $(\bm{x}^{\textsc{f}}_i, \bm{x}^{\textsc{f}}_j)$ and $(\bm{x}^{\textsc{f}}_i, \widetilde{\bm{x}}^{\textsc{f}}_j)$. 

\xhdr{Contrastive frequency loss} We calculate frequency-based contrastive loss for sample $\bm{x}_i$ as:
\begin{equation}
\label{eq:frequency_loss}
\mathcal{L}_{\textsc{f}, i} = d(\bm{h}^{\textsc{f}}_i, \widetilde{\bm{h}}^{\textsc{f}}_i, \mathcal{D}^{\textrm{pret}}) = -\textrm{log}\frac{\textrm{exp}(\textrm{sim}(\bm{h}^{\textsc{f}}_i, \widetilde{\bm{h}}^{\textsc{f}}_i)/\tau )}{\sum_{\bm{x}_j \in \mathcal{D}^{\textrm{pret}}} \mathbbm{1}_{i \neq j}\textrm{exp}(\textrm{sim}(\bm{h}^{\textsc{f}}_i,G_{\textsc{f}}(\bm{x}_j))/\tau )}.
\end{equation}
In preliminary experiments, we find that the value of $\tau$ has little effect on performance and use the same $\tau$ throughout all experiments.
The $\mathcal{L}_{\textsc{f}, i}$ yield a frequency encoder $G_{\textsc{f}}$ producing embeddings invariant to frequency spectrum perturbations. 


\subsection{Time-Frequency Consistency}
\label{sub:TF-C}
We develop a consistency loss item $\mathcal{L}_{\textsc{C},i}$ to urge the learned embeddings to satisfy TF-C: for a given sample, its time-based and frequency-based embeddings (and their local neighborhoods) are supposed to be close to each other (see Sec.~\ref{sec:problem_formulation} for justification). 
To make sure the distance between embeddings is measurable, we map $\bm{h}^{\textsc{t}}_i$ from time space and $\bm{h}^{\textsc{f}}_i$ from frequency space to a joint time-frequency space through projectors $R_{\textsc{t}}$ and $R_{\textsc{f}}$, respectively. In specific, for every input sample $\bm{x}_i$, we have four embeddings, which are $\bm{z}^{\textsc{t}}_i = R_{\textsc{t}}(\bm{h}^{\textsc{t}}_i)$, $\widetilde{\bm{z}}^{\textsc{t}}_i = R_{\textsc{t}}(\widetilde{\bm{h}}^{\textsc{t}}_i)$, $\bm{z}^{\textsc{f}}_i = R_{\textsc{f}}(\bm{h}^{\textsc{f}}_i)$, and $\widetilde{\bm{z}}^{\textsc{f}}_i = R_{\textsc{f}}(\widetilde{\bm{h}}^{\textsc{f}}_i)$. The first two embeddings are generated based on temporal characteristics and the latter two embeddings are produced based on the properties of frequency spectrum. 

To enforce the embeddings in the time-frequency space subject to TF-C, we design a consistency loss $\mathcal{L}_{\textsc{C}, i}$ that measures the distance between a time-based embedding and a frequency-based embedding.
We use
$S^{\textsc{t}\textsc{F}}_i = d(\bm{z}^{\textsc{t}}_i, \bm{z}^{\textsc{f}}_i, \mathcal{D}^{\textrm{pret}})$ 
to denote the distance between $\bm{z}^{\textsc{t}}_i$ and $\bm{z}^{\textsc{f}}_i)$. 
Similarly, we define $S^{\textsc{t}\widetilde{\textsc{f}}}_i$, $S^{\widetilde{\textsc{t}}\textsc{f}}_i$, and $S^{\widetilde{\textsc{t}}\widetilde{\textsc{f}}}_i$.
Note, in this time-frequency space, we don't consider the distance between  $\bm{z}^{\textsc{t}}_i$ and $\widetilde{\bm{z}}^{\textsc{t}}_i$ where the two embeddings are from the same domain (\ie, time domain). The same applies to pair the distance between $\bm{z}^{\textsc{f}}_i$ and $\widetilde{\bm{z}}^{\textsc{f}}_i$.
We have already considered information of above two pairs in the calculation of $\mathcal{L}_{\textsc{t}, i}$ and $\mathcal{L}_{\textsc{f}, i}$. 

Next, let's closely observe $S^{\textsc{t}\textsc{F}}_i$ and $S^{\textsc{t}\widetilde{\textsc{f}}}_i$ that involve three embeddings: $\bm{z}^{\textsc{t}}_i$, $\bm{z}^{\textsc{f}}_i$, and $\widetilde{\bm{z}}^{\textsc{f}}_i$. Here, $\bm{z}^{\textsc{t}}_i$ and $\bm{z}^{\textsc{f}}_i$ are learned from the original sample ($\bm{x}^{\textsc{t}}_i$ and $\bm{x}^{\textsc{f}}_i$) while $\widetilde{\bm{z}}^{\textsc{f}}_i$ is learned from the augmented $\widetilde{\bm{x}}^{\textsc{f}}_i$.
Thus, intuitively, $\bm{z}^{\textsc{t}}_i$ should be closer to $\bm{z}^{\textsc{f}}_i$ in comparison to $\widetilde{\bm{z}}^{\textsc{f}}_i$.
Motivated by the relative relationship, we encourage the proposed model to learn a $S^{\textsc{t}\textsc{F}}_i$ that is smaller than $S^{\textsc{t}\widetilde{\textsc{f}}}_i$. Inspired by the triplet loss~\cite{hoffer_deep_2014},
we design $(S^{\textsc{t}\textsc{F}}_i -S^{\textsc{t}\widetilde{\textsc{f}}}_i +\delta )$ as a term of consistency loss  $\mathcal{L}_{\textsc{c}, i}$ where $\delta$ is a given constant margin to keep negative samples far apart~\cite{balntas_learning_2016}. This term optimizes the model towards a smaller $S^{\textsc{t}\textsc{F}}_i$ and relatively larger $S^{\textsc{t}\widetilde{\textsc{f}}}_i$.
Similarly, $S^{\textsc{t}\textsc{f}}_i$ is supposed to be smaller than $S^{\widetilde{\textsc{t}}\textsc{f}}_i$ and $S^{\widetilde{\textsc{t}}\widetilde{\textsc{f}}}_i$. In summary, we calculate the consistency loss $\mathcal{L}_{\textsc{c}, i}$ for sample $\bm{x}_i$ by: 
\begin{equation}
\label{eq:loss_c}
\mathcal{L}_{\textsc{c}, i} = \sum_{S^{\textrm{pair}}} (S^{\textsc{t}\textsc{f}}_i - S^{\textrm{pair}}_i+\delta), \quad S^{\textrm{pair}} \in \{ S^{\textsc{t}\widetilde{\textsc{f}}}_i, S^{\widetilde{\textsc{t}}\textsc{f}}_i, S^{\widetilde{\textsc{t}}\widetilde{\textsc{f}}}_i\},
\end{equation}
where $S^{\textrm{pair}}_i$ denotes the distance between a time-based embedding (\eg, $\bm{z}^{\textsc{t}}_i$ or $\widetilde{\bm{z}}^{\textsc{t}}_i$) and a frequency-based embedding (\eg, $\bm{z}^{\textsc{f}}_i$ or $\widetilde{\bm{z}}^{\textsc{f}}_i$). In each pair, there is at least one embedding that is derived from augmented sample instead of the original sample. The $\delta$ is a pre-defined constant.
By combining all the triplet loss items, $\mathcal{L}_{\textsc{c}}$ encourages the pre-training model to capture the consistency between time-based and frequency-based embeddings in model optimization. 
Note, although the Eq.~\ref{eq:loss_c} does not explicitly measure the loss across different time series samples (\eg, $\bm{x}_i$ and $\bm{x}_j$), the cross-sample relationships are implicitly covered in the calculation of $S^{\textsc{t}\textsc{f}}_i$ and $S^{\textrm{pair}}_i$.









\subsection{Implementation and Technical Details}
The overall loss function in pre-training has three terms.
First, the time-based contrastive loss $\mathcal{L}_{\textsc{t}}$ urges the model to learn embeddings invariant to temporal augmentations. Second, the frequency-based contrastive loss $\mathcal{L}_{\textsc{f}}$ promotes learning of embeddings invariant to frequency spectrum-based augmentations. Third, the consistency loss $\mathcal{L}_{\textsc{c}}$ guides the model to retain the consistency between time-based and frequency-based embeddings.
In summary, the pre-training loss is defined as:
\begin{equation}
\label{eq:joint_loss}
\mathcal{L}_{\textrm{TF-C}, i} = \lambda(\mathcal{L}_{\textsc{t}, i} + \mathcal{L}_{\textsc{f}, i}) + (1- \lambda)\mathcal{L}_{\textsc{c}, i}
\end{equation}
where $\lambda$ controls the relative importance of the contrastive and consistency losses.
We calculate the total loss by summing $\mathcal{L}_{\textrm{TF-C}, i}$ across all pre-training samples. In implementation, the contrastive losses are calculated within the batch.
From our problem definition, the model $\mathcal{F}$ we want to learn is the combination of neural networks $G_{\textsc{t}}$, $R_{\textsc{t}}$, $G_{\textsc{f}}$, and $R_{\textsc{f}}$. 
When pre-training is completed, we store parameters of entire model, and denote it as $\mathcal{F}(\cdot, \Theta)$ where $\Theta$ represents all trainable parameters. 
%
When a sample $\bm{x}^{\textrm{tune}}_i$ is presented, fine-tuned model $\mathcal{F}$ generates an embedding $\bm{z}^{\textrm{tune}}_i$ via concatenation as: $\bm{z}^{\textrm{tune}}_i = \mathcal{F}(\bm{x}^{\textrm{tune}}_i, \Phi) = [\bm{z}^{\textrm{tune}, \textsc{t}}_i ; \bm{z}^{\textrm{tune}, \textsc{f}}_i]$ where $\Phi$ are fine-tuned model's parameters. 


\section{Experiments}\label{sec:experiments}

We compare the developed TF-C model with 10 baselines on 8 diverse datasets. We investigate the time series classification tasks in the context of one-to-one and one-to-many transfer learning setups (the many-to-one setting is fundamentally different as discussed in Appendix~\ref{app:many-to-one}). We also assess TF-C in extensive downstream tasks including clustering and anomaly detection.

\textbf{Datasets.} 
 (1) \textbf{\textsc{SleepEeg}} \cite{kemp_analysis_2000} has 371,055 univariate brainwaves (EEG; 100 Hz) collected from 197 individuals. Each sample is associated with one of five sleeping stages. (2) \textbf{\textsc{Epilepsy}} \cite{andrzejak_indications_2001} monitors the brain activities of 500 subjects  with single-channel EEG sensor (174 Hz). A sample is labeled in binary based on whether the subject \revise{has} epilepsy or not. (3) \textbf{\textsc{Fd-a}} \cite{lessmeier_condition_2016} gathers the vibration signals from rolling bearing from a mechanical system aiming at fault detection. Every sample has 5,120 timestamps and an indicator for one out of three mechanical device states. (4) \textbf{\textsc{Fd-b}} \cite{lessmeier_condition_2016} has the same setting as the \textbf{\textsc{Fd-a}} but the rolling bearings are performed in different working conditions (\eg, varying rotational speed). (5) \textbf{\textsc{Har}} \cite{anguita_public_2013} has 10,299 9-dimension samples from 6 daily activities. (6) \textbf{\textsc{Gesture}} 
\cite{liu_uwave_2009} includes 440 samples that are collected from 8 hand gestures recorded by an accelerometer. (7) \textbf{\textsc{Ecg}} 
\cite{clifford_af_2017} contains 8,528 single-sensor ECG recordings with sorted into four classes based on human physiology. (8) \textbf{\textsc{Emg}} \cite{goldberger_physiobank_2000} consists of 163 EMG samples with 3-class labels implying muscular diseases. Dataset labels are not used in pre-training. Further dataset statistics are in Appendix~\ref{app:4scenarios} and Table~\ref{tab:data_statistics}.



\textbf{Baselines.} We consider 10 baseline methods. This includes 8 state-of-the-art methods: TS-SD~\cite{shi_self-supervised_2021}, TS2vec~\cite{yue_ts2vec_2022}, CLOCS~\cite{kiyasseh_clocs_nodate}, Mixing-up~\cite{wickstrom_mixing_2022}, TS-TCC~\cite{eldele_time-series_2021}, SimCLR~\cite{tang_exploring_2021}, TNC~\cite{tonekaboni_unsupervised_2021}, and CPC~\cite{oord_representation_2018}. The TS2Vec, TS-TCC, SimCLR, TNC, and CPC are designed for representation learning on a single dataset rather than for transfer learning, so we apply them to fit our settings and make the results comparable. As the training of TNC and CPC are very time-consuming and relatively less competitive (Table~\ref{tab:sleep2epilepsy}), we only compare them in the one-to-one setting (scenario 1) while not in other experiments. To examine the utility of pre-training, we consider two additional approaches that are applied directly to fine-tuning datasets without any pre-training: Non-DL (a non-deep learning KNN model) and Random Init. (randomly initializes the fine-tuning model).
The evaluation metrics are accuracy, precision (macro-averaged), recall, F1 score, AUROC, and AUPRC.


\xhdr{Implementation} We use two 3-layer 1-D ResNets~\cite{ramanathan2021fall} as backbones for encoders $G_\textsc{t}$ and $G_\textsc{f}$. Our datasets contain long time series (samples in \textsc{Fd-a} and \textsc{Fd-b} have 5,120 observations), and preliminary experiments identified ResNet as a better option than a Transformer variant~\cite{zerveas2021transformer}. We use 2 fully-connected layers for $R_\textsc{t}$ and $R_\textsc{f}$, with no sharing of parameters. We set $E =1$ and $\alpha=0.5$ in frequency augmentations and $\tau = 0.2$, $\delta = 1$, $\lambda=0.5$ in loss functions. Reported are mean and standard deviation values across 5 independent runs (both pre-training and fine-tuning) on the same data split. Results for KNN (K=2) do not change so the standard deviation is zero. 
Method details and hyper-parameter selection are in Appendix~\ref{app:baseline_implementation}.

\subsection{Results: One-to-One Pre-Training Evaluation}
\label{sub:one_to_one}
\xhdr{Setup}
In one-to-one evaluation, we pre-train a model on \textit{one} pre-training dataset and use it for fine-tuning on \textit{one} target dataset only. 
\underline{Scenario 1} (\textsc{SleepEeg} $\rightarrow$ \textsc{Epilepsy}): Pre-training is done on \textsc{SleepEeg} and fine-tuning on \textsc{Epilepsy}. While both datasets describe a single-channel EEG, the signals are from different channels/positions on scalps, track different physiology (sleep vs.~epilepsy), and are collected from different patients. 
\underline{Scenario 2} (\textsc{Fd-a} $\rightarrow$ \textsc{Fd-b}): Datasets describe mechanical devices that operate in different working conditions, including rotational speed, load torque, and radial force.
\underline{Scenario 3} (\textsc{Har} $\rightarrow$ \textsc{Gesture}): Datasets record different activities (6 types of human daily activities vs.~8 hand gestures). While both datasets contain acceleration signals, \textsc{Har} has 9 channels while \textsc{Gesture} has 1 channel. 
\underline{Scenario 4} (\textsc{Ecg} $\rightarrow$ \textsc{Emg}): While both are physiological datasets, the \textsc{Ecg} records the electrical signal from the heart whereas \textsc{Emg} measures muscle response in response to a nerve's stimulation of the muscle. 
We note that the discrepancies between pre-training and fine-tuning datasets in the above four scenarios are substantial, and they cover a diverse range of variation in time series datasets: varying semantic meaning, sampling frequency, time series length, number of classes, and system factors (\eg, number of devices or subjects).
The setup is further challenged by the relatively small number of samples available for fine-tuning (\textsc{Epilepsy}: 60; \textsc{Fd-b}: 60; \textsc{Gesture}: 480; \textsc{Emg}: 122).
Further details are in Appendix~\ref{app:one_to_one_details}. 

\begin{table}[]
\centering
\def\arraystretch{1.0}
\caption{\textbf{One-to-one pre-training evaluation (Scenario 3).} Pre-training is performed on \textsc{Har}, followed by fine-tuning on \textsc{Gesture}. Results for other three scenarios are shown in Tables~\ref{tab:sleep2epilepsy}-\ref{app:ecg2emg}.
}
\label{tab:har2Gesture}
\resizebox{\textwidth}{!}{%
\begin{tabular}{Hllrrrrr}
\toprule
 & \textbf{Models} & \textbf{Accuracy} & \multicolumn{1}{l}{\textbf{Precision}} & \multicolumn{1}{l}{\textbf{Recall}} & \multicolumn{1}{l}{\textbf{F1 score}} & \multicolumn{1}{l}{\textbf{AUROC}} & \multicolumn{1}{l}{\textbf{AUPRC}} \\ \midrule
\multirow{9}{*}{\begin{tabular}[c]{@{}l@{}}SleepEDF\\ $\rightarrow$\\ \\ Epilepsy\end{tabular}} 
& \textbf{Non-DL (KNN)} & 0.6766\std{0.0000} & \multicolumn{1}{l}{0.6500\std{0.0000}} & \multicolumn{1}{l}{0.6821\std{0.0000}} & \multicolumn{1}{l}{0.6442\std{0.0000}} & \multicolumn{1}{l}{0.8190\std{0.0000}} & \multicolumn{1}{l}{0.5231\std{0.0000}} \\
 & \textbf{Random Init.} & 0.4219\std{0.0865} & \multicolumn{1}{l}{0.4751\std{0.0925}} & \multicolumn{1}{l}{0.4963\std{0.1026}} & \multicolumn{1}{l}{0.4886\std{0.0967}} & \multicolumn{1}{l}{0.7129\std{0.1206}} & \multicolumn{1}{l}{0.3358\std{0.1194}} \\ \cmidrule{2-8}
 & \textbf{TS-SD} & 0.6937\std{0.0533} & 0.6806\std{0.0496} & 0.6883\std{0.0525} & 0.6785\std{0.0495} & 0.8708\std{0.0305} & 0.6261\std{0.0790} \\
 & \textbf{TS2vec} & 0.6453\std{0.0260} & 0.6287\std{0.0339} & 0.6451\std{0.0218} & 0.6261\std{0.0294} & 0.8890\std{0.0054} & 0.6670\std{0.0118} \\
 & \textbf{CLOCS} & 0.4731\std{0.0229} & 0.4639\std{0.0432} & 0.4766\std{0.0266} & 0.4392\std{0.0198} & 0.8161\std{0.0068} & 0.4916\std{0.0103} \\
 & \textbf{Mixing-up} & 0.7183\std{0.0123} & 0.7001\std{0.0166} & 0.7183\std{0.0123} & 0.6991\std{0.0145} & \textbf{0.9127\std{0.0018}} & 0.7654\std{0.0071} \\
 & \textbf{TS-TCC} & 0.7593\std{0.0242} & 0.7668\std{0.0257} & 0.7566\std{0.0231} & 0.7457\std{0.0210} & 0.8866\std{0.0040} & 0.7217\std{0.0121} \\
 & \textbf{SimCLR} & 0.4383\std{0.0652} & 0.4255\std{0.1072} & 0.4383\std{0.0652} & 0.3713\std{0.0919} & 0.7721\std{0.0559} & 0.4116\std{0.0971} \\
  & \textbf{TF-C (Ours)} & \textbf{0.7824\std{0.0237}} & \textbf{0.7982\std{0.0496}} & \textbf{0.8011\std{0.0322}} & \textbf{0.7991\std{0.0296}} & 0.9052\std{0.0136} & \textbf{0.7861\std{0.0149}} \\
\bottomrule
\end{tabular}
}
\vspace{-5mm}
\end{table}

\xhdr{Results}
The results for the four scenarios are shown in Table~\ref{tab:har2Gesture} and Tables~\ref{tab:sleep2epilepsy}-\ref{app:ecg2emg}. Overall, our TF-C model has won 16 out of 24 tests (6 metrics in 4 scenarios) and is the second-best performer in only 8 other tests. 
We report all metrics but discuss the F1 score in the following. 
On average, our TF-C model claims a large margin of 15.4\% over all baselines. Although the strongest baseline is varying (such as TS-TCC in Scenario 2; Mixing-up in Scenario 3), our model outperforms the strongest baselines by 1.5\% across all scenarios. 
Specifically, as shown in Table~\ref{tab:har2Gesture} (\textsc{Har} $\rightarrow$ \textsc{Gesture}; Scenario 3),
TF-C achieves the highest performance of 79.91\% in F1 score, which yields a margin of 7.2\% over the best baseline TS-TCC (74.57\%). 
One potential explanation is that Scenario 3 involves a complex dataset (\textsc{Har} has 6 classes while \textsc{Gesture} has 8 classes) that can be difficult to model.
The complexity of Scenario 3 is further verified by poor performance of all models ($\pm 80\%$) relative to performance on other Scenarios ($\pm 90\%$): TF-C shows strong robustness by learning more generalizable representations. 
Additionally, we visualize the learned representations in time-frequency space (Appendix~\ref{app:visualization}), and the analyses provide further support for the TF-C property.



\begin{table}[]
\centering
\def\arraystretch{1.0}
\caption{\textbf{One-to-many pre-training evaluation.} Pre-training is performed on \textsc{SleepEeg}, followed by an independent fine-tuning on \textsc{Epilepsy}, \textsc{Fd-b}, \textsc{Gesture}, and \textsc{Emg}.}
\label{tab:one_to_many}
\resizebox{\textwidth}{!}{%
\begin{tabular}{cllrrrrr}
\toprule
 \textbf{Scenarios} & \textbf{Models} & \textbf{Accuracy} & \multicolumn{1}{l}{\textbf{Precision}} & \multicolumn{1}{l}{\textbf{Recall}} & \multicolumn{1}{l}{\textbf{F1 score}} & \multicolumn{1}{l}{\textbf{AUROC}} & \multicolumn{1}{l}{\textbf{AUPRC}} \\ \midrule
\multirow{9}{*}{\begin{tabular}[c]{@{}l@{}}\;\textsc{SleepEeg}\\ $\quad\;\;\; \downarrow$\\\;\textsc{Epilepsy}\end{tabular}} & \textbf{Non-DL (KNN)} & 0.8525\std{0.0000} & \multicolumn{1}{l}{0.8639\std{0.0000}} & \multicolumn{1}{l}{0.6431\std{0.0000}} & \multicolumn{1}{l}{0.6791\std{0.0000}} & \multicolumn{1}{l}{0.6434\std{0.0000}} & \multicolumn{1}{l}{0.6279\std{0.0000}} \\
 & \textbf{Random Init.} & 0.8983\std{0.0656} & \multicolumn{1}{l}{0.9213\std{0.1369}} & \multicolumn{1}{l}{0.7447\std{0.1135}} & \multicolumn{1}{l}{0.7959\std{0.1208}} & \multicolumn{1}{l}{0.8578\std{0.2153}} & \multicolumn{1}{l}{0.6489\std{0.1926}} \\ \cmidrule{2-8}
 & \textbf{TS-SD} & 0.8952\std{0.0522} & 0.8018\std{0.2244} & 0.7647\std{0.1485} & 0.7767\std{0.1855} & 0.7677\std{0.2452} & 0.7940\std{0.1825} \\
 & \textbf{TS2vec} & 0.9395\std{0.0044} & 0.9059\std{0.0116} & 0.9039\std{0.0118} & 0.9045\std{0.0067} & 0.9587\std{0.0086} & 0.9430\std{0.0103} \\
 & \textbf{CLOCS} & \textbf{0.9507\std{0.0027}} & 0.9301\std{0.0067} & \textbf{0.9127\std{0.0165}} & \textbf{0.9206\std{0.0066}} & 0.9803\std{0.0023} & 0.9609\std{0.0116} \\
 & \textbf{Mixing-up} & 0.8021\std{0.0000} & 0.4011\std{0.0000} & 0.5000\std{0.0000} & 0.4451\std{0.0000} & 0.9743\std{0.0081} & 0.9618\std{0.0104} \\
 & \textbf{TS-TCC} & 0.9253\std{0.0098} & 0.9451\std{0.0049} & 0.8181\std{0.0257} & 0.8633\std{0.0215} & 0.9842\std{0.0034} & 0.9744\std{0.0043} \\
 & \textbf{SimCLR} & 0.9071\std{0.0344} & 0.9221\std{0.0166} & 0.7864\std{0.1071} & 0.8178\std{0.0998} & 0.9045\std{0.0539} & 0.9128\std{0.0205} \\
 & \textbf{TF-C (Ours)} & 0.9495\std{0.0249} & \textbf{0.9456\std{0.0108}} & 0.8908\std{0.0216} & 0.9149\std{0.0534} & \textbf{0.9811\std{0.0237}} & \textbf{0.9703\std{0.0199}} \\ 
 \midrule
\multirow{9}{*}{\begin{tabular}[c]{@{}l@{}}\;\textsc{SleepEeg}\\ $\quad\;\;\; \downarrow$\\ \;\;\;\textsc{Fd-b}\end{tabular}} & \textbf{Non-DL (KNN)} & 0.4473\std{0.0000} & \multicolumn{1}{l}{0.2847\std{0.0000}} & \multicolumn{1}{l}{0.3275\std{0.0000}} & \multicolumn{1}{l}{0.2284\std{0.0000}} & \multicolumn{1}{l}{0.4946\std{0.0000}} & \multicolumn{1}{l}{0.3308\std{0.0000}} \\
 & \textbf{Random Init.} & 0.4736\std{0.0623} & \multicolumn{1}{l}{0.4829\std{0.0529}} & \multicolumn{1}{l}{0.5235\std{0.1023}} & \multicolumn{1}{l}{0.4911\std{0.0590}} & \multicolumn{1}{l}{0.7864\std{0.0349}} & \multicolumn{1}{l}{0.7528\std{0.0254}} \\ \cmidrule{2-8}
 & \textbf{TS-SD} & 0.5566\std{0.0210} & 0.5710\std{0.0535} & 0.6054\std{0.0272} & 0.5703\std{0.0328} & 0.7196\std{0.0113} & 0.5693\std{0.0532} \\
 & \textbf{TS2vec} & 0.4790\std{0.0113} & 0.4339\std{0.0092} & 0.4842\std{0.0197} & 0.4389\std{0.0107} & 0.6463\std{0.0130} & 0.4442\std{0.0162} \\
 & \textbf{CLOCS} & 0.4927\std{0.0310} & 0.4824\std{0.0316} & 0.5873\std{0.0387} & 0.4746\std{0.0485} & 0.6992\std{0.0099} & 0.5501\std{0.0365} \\
 & \textbf{Mixing-up} & 0.6789\std{0.0246} & 0.7146\std{0.0343} & \textbf{0.7613\std{0.0198}} & 0.7273\std{0.0228} & 0.8209\std{0.0035} & 0.7707\std{0.0042} \\
 & \textbf{TS-TCC} & 0.5499\std{0.0220} & 0.5279\std{0.0293} & 0.6396\std{0.0178} & 0.5418\std{0.0338} & 0.7329\std{0.0203} & 0.5824\std{0.0468} \\
 & \textbf{SimCLR} & 0.4917\std{0.0437} & 0.5446\std{0.1024} & 0.4760\std{0.0885} & 0.4224\std{0.1138} & 0.6619\std{0.0219} & 0.5009\std{0.0477} \\
 & \textbf{TF-C (Ours)} & \textbf{0.6938\std{0.0231}} & \multicolumn{1}{l}{\textbf{0.7559\std{0.0349}}} & \multicolumn{1}{l}{0.7202\std{0.0257}} & \multicolumn{1}{l}{\textbf{0.7487\std{0.0268}}} & \multicolumn{1}{l}{\textbf{0.8965\std{0.0135}}} & \multicolumn{1}{l}{\textbf{0.7871\std{0.0267}}} \\ \midrule 
\multirow{9}{*}{\begin{tabular}[c]{@{}l@{}}\;\textsc{SleepEeg}\\ $\quad\;\;\;\; \downarrow$\\ \;\;\textsc{Gesture}\end{tabular}} & \textbf{Non-DL (KNN)} & 0.6833\std{0.0000} & \multicolumn{1}{l}{0.6501\std{0.0000}} & \multicolumn{1}{l}{0.6833\std{0.0000}} & \multicolumn{1}{l}{0.6443\std{0.0000}} & \multicolumn{1}{l}{0.8190\std{0.0000}} & \multicolumn{1}{l}{0.5232\std{0.0000}} \\
 & \textbf{Random Init.} & 0.4219\std{0.0629} & \multicolumn{1}{l}{0.4751\std{0.0175}} & \multicolumn{1}{l}{0.4963\std{0.0679}} & \multicolumn{1}{l}{0.4886\std{0.0459}} & \multicolumn{1}{l}{0.7129\std{0.0166}} & \multicolumn{1}{l}{0.3358\std{0.1439}} \\ \cmidrule{2-8}
 & \textbf{TS-SD} & 0.6922\std{0.0444} & 0.6698\std{0.0472} & 0.6867\std{0.0488} & 0.6656\std{0.0443} & 0.8725\std{0.0324} & 0.6185\std{0.0966} \\
 & \textbf{TS2vec} & 0.6917\std{0.0333} & 0.6545\std{0.0358} & 0.6854\std{0.0349} & 0.6570\std{0.0392} & 0.8968\std{0.0123} & 0.6989\std{0.0346} \\
 & \textbf{CLOCS} & 0.4433\std{0.0518} & 0.4237\std{0.0794} & 0.4433\std{0.0518} & 0.4014\std{0.0602} & 0.8073\std{0.0109} & 0.4460\std{0.0384} \\
 & \textbf{Mixing-up} & 0.6933\std{0.0231} & 0.6719\std{0.0232} & 0.6933\std{0.0231} & 0.6497\std{0.0306} & 0.8915\std{0.0261} & 0.7279\std{0.0558} \\
 & \textbf{TS-TCC} & 0.7188\std{0.0349} & 0.7135\std{0.0352} & 0.7167\std{0.0373} & 0.6984\std{0.0360} & 0.9099\std{0.0085} & 0.7675\std{0.0201} \\
 & \textbf{SimCLR} & 0.4804\std{0.0594} & 0.5946\std{0.1623} & 0.5411\std{0.1946} & 0.4955\std{0.1870} & 0.8131\std{0.0521} & 0.5076\std{0.1588} \\
 & \textbf{TF-C (Ours)} & \textbf{0.7642\std{0.0196}} & \multicolumn{1}{l}{\textbf{0.7731\std{0.0355}}} & \multicolumn{1}{l}{\textbf{0.7429\std{0.0268}}} & \multicolumn{1}{l}{\textbf{0.7572\std{0.0311}}} & \multicolumn{1}{l}{\textbf{0.9238\std{0.0159}}} & \multicolumn{1}{l}{\textbf{0.7961\std{0.0109}}} \\ \midrule 
\multirow{9}{*}{\begin{tabular}[c]{@{}l@{}}\;\textsc{SleepEeg}\\ $\quad\;\;\; \downarrow$\\ \;\;\;\;\textsc{Emg}\end{tabular}} & \textbf{Non-DL (KNN)} & 0.4390\std{0.0000} & \multicolumn{1}{l}{0.3772\std{0.0000}} & \multicolumn{1}{l}{0.5143\std{0.0000}} & \multicolumn{1}{l}{0.3979\std{0.0000}} & \multicolumn{1}{l}{0.6025\std{0.0000}} & \multicolumn{1}{l}{0.4084\std{0.0000}} \\
 & \textbf{Random Init.} & 0.7780\std{0.0729} & \multicolumn{1}{l}{0.5909\std{0.0625}} & \multicolumn{1}{l}{0.6667\std{0.0135}} & \multicolumn{1}{l}{0.6238\std{0.0267}} & \multicolumn{1}{l}{0.9109\std{0.1239}} & \multicolumn{1}{l}{0.7771\std{0.1427}} \\ \cmidrule{2-8}
 & \textbf{TS-SD} & 0.4606\std{0.0000} & \multicolumn{1}{l}{0.1545\std{0.0000}} & 0.3333\std{0.0000} & 0.2111\std{0.0000} & 0.5005\std{0.0126} & 0.3775\std{0.0110} \\
 & \textbf{TS2vec} & 0.7854\std{0.0318} & \multicolumn{1}{l}{\textbf{0.8040\std{0.0750}}} & 0.6785\std{0.0396} & 0.6766\std{0.0501} & \textbf{0.9331\std{0.0164}} & \textbf{0.8436\std{0.0372}} \\
 & \textbf{CLOCS} & 0.6985\std{0.0323} & \multicolumn{1}{l}{0.5306\std{0.0750}} & 0.5354\std{0.0291} & 0.5139\std{0.0409} & 0.7923\std{0.0573} & 0.6484\std{0.0680} \\
 & \textbf{Mixing-up} & 0.3024\std{0.0534} & \multicolumn{1}{l}{0.1099\std{0.0126}} & 0.2583\std{0.0456} & 0.1541\std{0.0204} & 0.4506\std{0.1718} & 0.3660\std{0.1635} \\
 & \textbf{TS-TCC} & 0.7889\std{0.0192} & \multicolumn{1}{l}{0.5851\std{0.0974}} & 0.6310\std{0.0991} & 0.5904\std{0.0952} & 0.8851\std{0.0113} & 0.7939\std{0.0386} \\
 & \textbf{SimCLR} & 0.6146\std{0.0582} & \multicolumn{1}{l}{0.5361\std{0.1724}} & 0.4990\std{0.1214} & 0.4708\std{0.1486} & 0.7799\std{0.1344} & 0.6392\std{0.1596} \\
 & \textbf{TF-C (Ours)} & \textbf{0.8171\std{0.0287}} & \multicolumn{1}{l}{0.7265\std{0.0353}} & \textbf{0.8159\std{0.0289}} & \textbf{0.7683\std{0.0311}} & 0.9152\std{0.0211} & 0.8329\std{0.0137} \\ 
 \bottomrule 
\end{tabular}
}
\vspace{-6mm}
\end{table}

\subsection{Results: One-to-Many Pre-Training Evaluation}
\label{sub:one_to_many}
\xhdr{Setup} 
In one-to-many evaluation, pre-training is done using \textit{one} dataset followed by fine-tuning on \textit{multiple} target datasets independently without starting pre-training from scratch. Out of eight datasets, \textsc{SleepEeg} has most complex temporal dynamics~\cite{zhang2021deep} and is the largest (371,055 samples). For that reason, we pre-train a model on \textsc{SleepEeg} and separately fine-tune a well-pre-trained model on \textsc{Epilepsy}, \textsc{Fd-b}, \textsc{Gesture}, and \textsc{Emg}. 


\xhdr{Results} 
Results are shown in Table~\ref{tab:one_to_many}.
As there are fewer commonalities between EEG signals vs.~vibration, and acceleration vs.~EMG, we expect that transfer learning will be less effective for them than one-to-one evaluations.
The pre-training and fine-tuning datasets are largely different in the bottom three blocks (\textsc{SleepEeg} $\rightarrow$ \{\textsc{Fd-b}, \textsc{Gesture}, \textsc{Emg}\}). The large gap reasonably leads to a deterioration in baseline performances, however, our model has a noticeably higher tolerance to knowledge transfer across datasets with large gaps. 
Notably, We find that the proposed model with TF-C earned the best performance in 14 out of 18 settings in the three challenging settings: indicating our TF-C assumption is universal in time series. For example, our approach outperforms the strongest baseline by 8.4\% (in precision) when fine-tuning on \textsc{Gesture}. Our model has great potential to serve as a universal model when there is no large pre-training dataset that is similar to the small fine-tuning dataset.
Furthermore, the TF-C consistently outperforms KNN and Random Init. (which are not pre-trained) by a large margin of 42.8\% and 25.1\% (both in F1 score) on average.

\xhdr{Ablation study}
We evaluate how relevant the model components are for effective pre-training.
As shown in Table~\ref{tab:ablation} (\textsc{SleepEeg} $\rightarrow$ \textsc{Gesture}; Appendix~\ref{app:ablation}), removing $\mathcal{L}_{\textsc{C}}$, $\mathcal{L}_{\textsc{T}}$, and $\mathcal{L}_{\textsc{F}}$ result in performance degradation (precision) of 6.1\%, 7.2\%, and 6.7\%, respectively. 
To validate that the performance increment is not solely brought by a third loss term no matter what consistency it measures, we replaced consistency loss $\mathcal{L}_{\textsc{C}}$ with a loss term measuring the consistency within time space (named $\mathcal{L}_{\textsc{TT-C}}$) or within frequency space (named $\mathcal{L}_{\textsc{FF-C}}$). 
Results show our consistency loss outperforms $\mathcal{L}_{\textsc{TT-C}}$ and $\mathcal{L}_{\textsc{FF-C}}$
by 5.3\% and 7.2\% (accuracy), respectively.

\subsection{Additional Downstream Tasks: Clustering and Anomaly Detection}
\label{sub:clustering_anomaly}

\xhdr{Clustering Task}
%
We evaluate the clustering performance of TF-C taking \textsc{SleepEeg} $\rightarrow$ \textsc{Epilepsy} as an example.
Specifically, we added a K-means (K=2), as Epilepsy has 2 classes, on top of $z_i^{\textrm{tune}}$ in fine-tuning. We adopt commonly used evaluation metrics: Silhouette score, Adjusted Rand Index (ARI), and Normalized Mutual Information (NMI). 
Table~\ref{tab:clustering} shows our TF-C obtains the best clustering surpassing the strongest baseline (TS-TCC) by a large margin (5.4\% in Silhouette score). It conveys that TF-C can capture more distinctive representations with the knowledge transferred from pre-training, which is consistent with the superiority of TF-C in the above classification tasks.

\xhdr{Anomaly Detection Task}
We assess how TF-C performs on a sample-level anomaly detection task. 
Note we work on the sample-level rather than the observation-level anomaly detection. 
Based on global patterns, the former aims to detect abnormal time series samples instead of outlier observations in a sample (as in BTSF~\cite{yang2022unsupervised} and USAD~\cite{audibert2020usad}) which emphasizes local context. 
Specifically, 
In the scenario of \textsc{Fd-a} $\rightarrow$ \textsc{Fd-b}, we built a small subset of \textsc{Fd-b} with 1,000 samples, of which 900 are from undamaged bearings, and the remaining 100 are from bearings with inner or outer damage. Undamaged samples are considered “normal,” and inner/outer damaged samples are “outliers.” In fine-tuning, we used one-class SVM on top of learned representations $z_i^{\textrm{tune}}$. 
The experimental results (Table~\ref{tab:anomaly}) show that our TF-C outperforms five competitive baselines with 4.5\% in F-1 Score. Results show that the proposed TF-C is more sensitive to anomalous samples and can effectively detect the abnormal status in mechanical devices.

\section{Conclusion}\label{sec:conclusion}
We develop a pre-training approach that introduces time-frequency consistency (TF-C) as a mechanism to support knowledge transfer between time-series datasets. The approach uses self-supervised contrastive estimation and injects TF-C into pre-training, bringing time-based and frequency-based representations and their local neighborhoods close together in the latent space.

\xhdr{Limitations and future directions}
TF-C property can serve as a universal property for pre-training on diverse time series datasets. Additional generalizable properties, such as temporal autoregressive processes, could also be helpful for pre-training on time series.
Further, while our method expects as input a regularly sampled time series, it can handle irregularly sampled time series by using an encoder (such as Raindrop~\cite{zhang2021graph} and SeFT~\cite{horn2020set}) that can embed irregular time series. For frequency encoder inputs $\bm{x}^{\textsc{f}}_i$, alternatives include resampling or interpolation to obtain regularly sampled signals and using regular or non-uniform FFT operations.
Furthermore, TF-C's current embedding strategy and loss functions are favorable for classification, leveraging global information over tasks that use local context (\eg, forecasting). Results show that the TF-C approach performs well across broad downstream tasks, including classification, clustering, and anomaly detection (Sec.~\ref{sub:clustering_anomaly}).





\begin{ack}
We gratefully acknowledge support by US Air Force Contract No.~FA8702-15-D-0001, Harvard Data Science Initiative, and awards from Amazon Research, Bayer Early Excellence in Science, AstraZeneca Research, and Roche Alliance with Distinguished Scientists. T.T. is supported by the Under Secretary of Defense for Research and Engineering under US Air Force Contract No.~FA8702-15-D-0001. 
Any opinions, findings, conclusions or recommendations expressed in this material are those of the authors and do not necessarily reflect the views of the funders. 
\end{ack}

\newpage
\bibliographystyle{unsrt}  
\bibliography{updated_refs}

\clearpage
\section*{Broader Impacts}\label{sec:broader_impact}
Our approach for self-supervised pre-training improves classification performance on target datasets in different application scenarios. The recognition of time-frequency consistency as a universal property specific to time series data is a weak assumption that enables effective, task- and domain-agnostic transfer learning. We believe our work will inspire the research community to uncover other universal properties for transfer learning. We also hope our work will also attract more researchers to the more general problem of time series representation learning which is still underappreciated relative to problems from CV and NLP fields.

On the society level, our work, along the line of transfer learning, can facilitate more efficient use of time series data in various settings. For example, in medical settings, some diseases of clinical interest may have very small labelled dataset.
In this case, unlabelled data from patients of different diseases but with similar underlying physiological conditions can be used to pre-train the model. However, practitioners need to be aware of the limitations of the model, including that it may make biased predictions. Specifically, bias may exist in the source dataset used for pre-training due to an imbalance of samples from subjects of different demographic attributes. Also, the standardized medical protocols for collecting these datasets might be unsuitable for subjects with certain physiological attributes, creating unforeseen bias that may be transferred to fine-tuning. 


All datasets in this paper are publicly available and are not associated with any privacy or security concern. Furthermore, we have followed guidelines on responsible use specified by primary authors of the datasets used in the current work.

\section*{Checklist}



\begin{enumerate}

\item For all authors...
\begin{enumerate}
  \item Do the main claims made in the abstract and introduction accurately reflect the paper's contributions and scope?
    \answerYes{} In abstract and introduction, we claim that TF-C is a generalizable property of time series that can support pre-training, which is well-justified in Sec.~\ref{sec:problem_formulation} and experimentally demonstrated in Sec.~\ref{sec:experiments} (our model consistently performs comparatively to or above baseline methods). 
  \item Did you describe the limitations of your work?
    \answerYes{} See Section~\ref{sec:conclusion}.
  \item Did you discuss any potential negative societal impacts of your work?
    \answerYes{} See Broader Impact on Page 10. 
  \item Have you read the ethics review guidelines and ensured that your paper conforms to them?
    \answerYes{}
\end{enumerate}

\item If you are including theoretical results...
\begin{enumerate}
  \item Did you state the full set of assumptions of all theoretical results?
    \answerNA{}
        \item Did you include complete proofs of all theoretical results?
    \answerNA{}
\end{enumerate}

\item If you ran experiments...
\begin{enumerate}
  \item Did you include the code, data, and instructions needed to reproduce the main experimental results (either in the supplemental material or as a URL)?
    \answerYes{} Yes, we include an anonymous link (see Abstract) that provides the source codes with all implementation details, implementation of baselines, and eight datasets. The link will be updated to an non-anonymous link after acceptance.
  \item Did you specify all the training details (e.g., data splits, hyper-parameters, how they were chosen)?
    \answerYes{} See implementation details in Sec.~\ref{sec:experiments}. See Appendix~\ref{app:baseline_implementation} for baseline architectures and hyper-parameter settings. More details can be found in the included URL.
        \item Did you report error bars (e.g., with respect to the random seed after running experiments multiple times)?
    \answerYes{} We run experiments for 5 times and report the average value with standard deviation. See Table~\ref{tab:har2Gesture}, Tables~\ref{tab:sleep2epilepsy}-\ref{app:ecg2emg}, and Table~\ref{tab:one_to_many}.
        \item Did you include the total amount of compute and the type of resources used (e.g., type of GPUs, internal cluster, or cloud provider)?
    \answerYes{} See Appendix~\ref{app:baseline_implementation}.
\end{enumerate}

\item If you are using existing assets (e.g., code, data, models) or curating/releasing new assets...
\begin{enumerate}
  \item If your work uses existing assets, did you cite the creators?
    \answerYes{} We used eight existing datasets and 6 state-of-the-art baselines in contrastive learning and pre-training for time series. We cited the creators for every exist asset we used. See Sec.~\ref{sec:experiments}.
  \item Did you mention the license of the assets?
    \answerYes{} All dataset licenses are mentioned in the Appendix~\ref{app:4scenarios}.
  \item Did you include any new assets either in the supplemental material or as a URL?
    \answerYes{} See the anonymous link in Abstract.
  \item Did you discuss whether and how consent was obtained from people whose data you're using/curating?
    \answerNo{} All data we use is freely available for download, without any requirement to re-contact the data curator.
  \item Did you discuss whether the data you are using/curating contains personally identifiable information or offensive content?
    \answerNo{} Our datasets are public, well-established, and do not contain PII or offensive content
\end{enumerate}

\item If you used crowdsourcing or conducted research with human subjects...
\begin{enumerate}
  \item Did you include the full text of instructions given to participants and screenshots, if applicable?
    \answerNA{}
  \item Did you describe any potential participant risks, with links to Institutional Review Board (IRB) approvals, if applicable?
    \answerNA{}
  \item Did you include the estimated hourly wage paid to participants and the total amount spent on participant compensation?
    \answerNA{}
\end{enumerate}

\end{enumerate}

\clearpage

\appendix
\setcounter{page}{1}
\begin{appendices}

\section{Further information on the relationship between our pre-training approach and domain adaptation}
\label{app:difference_domain_adaptation}
Here we note our problem definition of pre-training is fundamentally different from domain adaptation~\citeS{wilson2020survey,zhou2020xhar,ott2022domain,da2020remaining,hao2022multi,zhang2021domain}\footnote{The supplementary document contains additional references, prefixed by `S`. These additional references are listed at the end of the Appendix.} in order to prevent any confusion between this work and domain adaptation methods. 
DA applies a model trained on a pre-training dataset (\ie, source dataset) to a different target dataset~\cite{singh2021clda,berthelot2021adamatch}. In contrast, self-supervised pre-training has four key differences with domain adaptation. (1) First, our model only requires the pre-training dataset while domain adaptation techniques generally require access to the target dataset~\citeS{wang2019hierarchical,tuia2021recent,liu2021optimal,lin2022cycda,xiao2021unsupervised,kim2021adaptive,zhang2021universal,zhu2021cross,chen2021self,fujii2021generative}. (2) Second, our model can be applied to multiple unseen target datasets (without re-training the pre-trained model for every target dataset) while domain adaptation approaches use the  target dataset during model training, \eg, \citeS{guan2021domain,mao2022new,xia2021adaptive,awais2021adversarial,kim2021adaptive,zhu2021cross,ma2021self} (see also Sec.~\ref{sub:one_to_many} for experimental results). (3) Third, our approach can be used in scenarios where the feature space in pre-training is different from that in the target dataset (see Scenarios 1, 3, and 4 in Sec.~\ref{sub:one_to_one} for experimental results). In contrast, domain adaptation methods usually restrict pre-training and target datasets to have the same feature space (but possible different distributions), \eg, \citeS{liu2021adversarial,mao2022new,xia2021adaptive,awais2021adversarial,zhang2021universal}. 


In summary, to support transfer learning across different time series datasets, a pre-training approach needs a capability to capture a generalizable property of time series, one that is shared across different time series datasets regardless of the specific semantic meaning of a time series signal (\eg, ECG, EMG, acceleration, vibration), conditions of data acquisition (\eg, variation across subjects and devices), sampling frequencies, etc. This work develops a self-supervised contrastive pre-training strategy that fulfills these requirements by injecting an appropriate inductive bias (called Time-Frequency Consistency, TF-C, into the model (Sec.~\ref{sec:problem_formulation}). 

Further, we clarify that the term `self-supervised' has different meanings in DA and in pre-training~\citeS{yuan2020self,reed2022self,chen2020adversarial,goyal2021self}.
The `self-supervised domain adaptation'~\citeS{akiva2021h2o,fujii2021generative,ma2021self,chen2021self} or `unsupervised domain adaptation'~\citeS{wilson2020survey,liu2021adversarial,wu2021dannet,xiao2021unsupervised,zhu2021cross} means that there are no labels in the target dataset, however that still requires labels in the pre-training dataset. In contrast, `self-supervised pre-training'~\citeS{baevski2019effectiveness,ragab2021self,hu2020strategies} (\ie, the problem studied here, in line with a breadth of existing literature on pre-training) indicates the setting where no labels are available in pre-training.

\section{Detailed differences with CoST and BTSF}
\label{app:BTSF}
Up to the submission of this manuscript, there is no existing contrastive augmentations in time series' frequency domain. There are two models, CoST~\cite{woo_cost_2022} and BTSF~\cite{yang2022unsupervised}, that involved frequency domain in contrastive learning, however, the proposed TF-C is fundamentally different with them in the following aspects. We take BTSF as an example while the differences also apply to CoST. 
\begin{itemize}
    \item Problem definitions for both papers are different. Our method is designed to produce generalizable representations that can transfer to a different time series dataset (going from pre-training to a fine-tuning dataset) for the purpose of transfer learning. In contrast, BTSF attempts to learn embeddings within the same dataset for the purpose of representation learning. Our model captures the TF-C property invariant to different time series (in terms of various temporal dynamics, semantic meaning, etc.) and can thus serve as a vehicle for transfer learning. In contrast, BTSF learns embeddings invariant to perturbations (i.e., instance-level dropout) of the same time series.
    \item The modeling of the frequency domain is different in both papers. We developed augmentations in the frequency domain based on the spectral properties of time series. In contrast, although BTSF involves a frequency domain, its data transformation is solely implemented in the time domain (using instance-level dropout; Sec. 3.1 in BTSF). That is, the BTSF method applies the FFT after augmenting samples in the time domain which can lead to information loss.
    \item BTSF emphasizes fusing temporal and spectral features to generate discriminative embeddings. Unlike BTSF, our model leverages the consistency between time-based and frequency-based embeddings to produce generalizable time series representations. Our model maps every sample to a time-frequency embedding space and constrains the relative relationships between embeddings (through triplet loss) according to the TF-C property. The underlying consistency allows TF-C to realize transfer learning across time series datasets, which is TF-C’s unique advantage.
\end{itemize}

In summary, TF-C introduces frequency domain augmentations in the sense that it directly perturbs the frequency spectrum. TF-C is the first method that uses frequency domain augmentations to enable transfer learning in time series.

\section{Time series invariance between time and frequency domains}
\label{app:invariance}
Here, we provide an analogy, from images to time series, to aid understanding of the `invariance' property of data representations.

In computer vision, it is well known that an image and its augmented views obtained by simple transformations (e.g., rotation, translation, scaling, etc.) can be used in different frameworks such as transformation prediction and contrastive instance discrimination to obtain invariant representations. These transformations are used as augmentations to guide the learning of self-supervised representations. For transformation prediction, the model learns representations that are equivariant to the selected transformation as the information embedded in the transformation needs to be embedded in the representation for the final layer to solve the pretext task. For contrastive instance discrimination, the representations are sensitive to the instances while learning invariances to transformations or views. The intuition behind such transformations is that the underlying object in the image is the same no matter how the image is rotated, translated, re-scaled, etc. In other words, the information carried by the original image and the transformed image is the same.

Similarly, the time domain and frequency domain representations carry the same information in time series. Thus it is reasonable to expect that by exploring local neighborhoods in the time domain and frequency domain and enforcing consistency of feature representations inter- and intra- domains, invariance properties are captured in our model via self-supervised learning. Thus, information carried in time and frequency domains of the same or similar time series sample should be the same. This invariance, formalized as `Time-Frequency Consistency', is helpful for self-supervised pre-training. We will include the above discussion in the camera-ready version.

\section{Additional information on datasets and pre-training evaluation}
\label{app:4scenarios}

\subsection{Datasets}
We use eight diverse time series datasets to evaluate our model. The datasets used in one-to-one and one-to-many pre-training evaluations are the same. The dataset statistics are shown in Table~\ref{tab:data_statistics}. Processed model-ready datasets are in our GitHub Repository (\url{https://anonymous.4open.science/r/TFC-pre-training-6B07}). Following is a detailed description of datasets. 

\begin{table}[t]
\centering
\caption{Description of datasets use in the four different application scenarios. We have two datasets (a pre-training dataset and a fine-tuning dataset) in each scenario. For the number of samples in fine-tuning dataset, "A/B/C" denotes we use A samples for fine-tuning, B samples for validation, and C samples for test. To test our effectiveness on small datasets (which is practically meaningful), we limit the fine-tuning set to a very small set with less than 320 samples. We ensure the fine-tuning set is balanced in terms of classes.
}
\label{tab:data_statistics}
\resizebox{\textwidth}{!}{%
\begin{tabular}{cccccccc}
\toprule
\multicolumn{1}{l}{Scenario \#} &  &\textbf{Dataset} & \multicolumn{1}{l}{\# Samples } & \multicolumn{1}{l}{\# Channels} & \multicolumn{1}{l}{\# Classes} & \multicolumn{1}{l}{Length} &  \multicolumn{1}{l}{Freq (Hz)}\\
\midrule
         1&Pre-training &\textbf{\textsc{SleepEeg}} &371,055 &1 &5 &200 &100\\
         &Fine-tuning &\textbf{\textsc{Epilepsy}} &60/20/11,420 &1 &2 &178 &174 \\
\midrule
         2&Pre-training &\textbf{\textsc{Fd-a}} &8,184 &1 &3 & 5,120 &64K\\
         &Fine-tuning &\textbf{\textsc{Fd-b}} &60/21/13,559 &1 &3 & 5,120 &64K\\
\midrule
         3&Pre-training &\textbf{\textsc{Har}} &10,299 &9 &6 &128 &50 \\
          &Fine-tuning &\textbf{\textsc{Gesture}} &320/120/120 & 3 &8 &315 & 100 \\
\midrule
         4&Pre-training &\textbf{\textsc{Ecg}}  &43,673 &1 &4 &1,500 &300 \\
          &Fine-tuning &\textbf{\textsc{Emg}} &122/41/41 &1 &3 & 1,500 & 4,000  \\ \bottomrule
\end{tabular}
}
\end{table}


\textbf{\textsc{SleepEeg}}~\cite{kemp_analysis_2000}. The dataset contains 153 whole-night sleeping electroencephalography (EEG) recordings produced by a sleep cassette. Data are collected from 82 healthy subjects. The 1-lead EEG signal is sampled at 100 Hz. We segment the EEG signals into segments (window size is 200) without overlapping, and each segment forms a sample. Every sample is associated with one of the five sleeping patterns/stages: Wake (W), Non-rapid eye movement (N1, N2, N3), and Rapid Eye Movement (REM). After segmentation, we have 371,055 EEG samples. The raw dataset (\url{https://www.physionet.org/content/sleep-edfx/1.0.0/}) is distributed under the Open Data Commons Attribution License v1.0.



\textbf{\textsc{Epilepsy}}~\cite{andrzejak_indications_2001}. The dataset contains single-channel EEG measurements from 500 subjects. For every subject, the brain activity was recorded for 23.6 seconds. The dataset was then divided and shuffled (to mitigate sample-subject association) into 11,500 samples of 1 second each, sampled at 178 Hz. The raw dataset features five classification labels corresponding to different states of subjects or measurement locations --- eyes open, eyes closed, EEG measured in the healthy brain region, EEG measured in the tumor region, and whether the subject has a seizure episode. To emphasize the distinction between positive and negative samples in terms of epilepsy, We merge the first four classes into one, and each time series sample has a binary label describing if the associated subject is experiencing a seizure or not. There are 11,500 EEG samples in total. To evaluate the performance of a pre-trained model on a small fine-tuning dataset, we choose a small set (60 samples; 30 samples for each class) for fine-tuning and assess the model with a validation set (20 samples; 10 samples for each class). Finally, the model with the best validation performance is used to make predictions on the test set (\ie, the remaining 11,420 samples). Statistics of fine-tuning, validation, and test sets in the other three target datasets are in Appendix~\ref{tab:data_statistics}.
The raw dataset (\url{https://repositori.upf.edu/handle/10230/42894}) is distributed under the Creative Commons License (CC-BY) 4.0.  

\textbf{\textsc{Fd-a}} and \textbf{\textsc{Fd-b}} ~\cite{lessmeier_condition_2016}. The dataset is generated by an electromechanical drive system that monitors the condition of rolling bearings and detects their failures. Four subsets of data are collected under various conditions, whose parameters include rotational speed, load torque, and radial force. Each rolling bearing can be undamaged, inner damaged, and outer damaged, which leads to three classes in total.
We denote the subsets corresponding to condition A and condition B as Faulty Detection Condition A (\textbf{\textsc{Fd-a}}) and Faulty Detection Condition B (\textbf{\textsc{Fd-b}}), respectively. 
Each original recording has a single channel with a sampling frequency of 64k Hz and lasts 4 seconds. To deal with lengthy recordings, we follow the procedure described by Eldele \etal~\cite{eldele_time-series_2021}. Specifically, we use a sliding window length of 5,120 observations and a shifting length of 1,024 or 4,096 to ensure that samples are relatively balanced between classes. The raw dataset (\url{https://mb.uni-paderborn.de/en/kat/main-research/datacenter/bearing-datacenter/data-sets-and-download}) is distributed under the Creative Commons Attribution-Non Commercial 4.0 International License.

\textbf{\textsc{Har}}~\cite{anguita_public_2013}. This dataset contains recordings of 30 health volunteers performing daily activities, including walking, walking upstairs, walking downstairs, sitting, standing, and lying. Prediction labels are the six activities. The wearable sensors on a smartphone measure triaxial linear acceleration and triaxial angular velocity at 50 Hz. 
After preprocessing and isolating gravitational acceleration from body acceleration, there are nine channels (\ie, 3-axis accelerometer, 3-axis gyroscope, and 3-axis magnetometer) in total. The raw dataset (\url{https://archive.ics.uci.edu/ml/datasets/Human+Activity+Recognition+Using+Smartphones}) is distributed as-is.
Any commercial use is not allowed.

\textbf{\textsc{Gesture}}~\cite{liu_uwave_2009}. The dataset contains accelerometer measurements of eight simple gestures that differ based on the paths of hand movement. The eight gestures are: hand swiping left, right, up, and down, hand waving in a counterclockwise or clockwise circle, hand waving in a square, and waving a right arrow. The classification labels are these eight different kinds of gestures.
The original paper reports the inclusion of 4,480 gesture samples, but through the UCR database, we can only recover 440 samples. 
The dataset is balanced, with 55 samples in each class, and is of an appropriate size for the fine-tuning. The original paper does not explicitly report sampling frequency but is presumably 100 Hz. The dataset uses three channels corresponding to three coordinate directions of acceleration. Note, when transferring knowledge from \textbf{\textsc{Har}} (nine channels) to \textbf{\textsc{Gesture}} (three channels): we use all the nine channels in pre-training if the model (\ie, TF-C, CLOCS, SimCLR, and TS-TCC) has channel generalization ability (\ie, allows different channel numbers in pre-training and fine-tuning datasets); otherwise, if the model (\ie, KNN, TS-SD, Mixing-up, and TS2vec) don't have channel generalization ability, we only use three acceleration channels in \textbf{\textsc{Har}} for pre-training.
%
The raw dataset is accessible through \url{http://www.timeseriesclassification.com/description.php?Dataset=UWaveGestureLibrary}. While the distribution license is not explicitly mentioned, the dataset is a public resource based on \cite{liu_uwave_2009}.



\textbf{\textsc{Ecg}}~\cite{clifford_af_2017}. This is the 2017 PhysioNet Challenge focusing on classifying ECG recordings. The single-lead ECG measures four different underlying conditions of cardiac arrhythmias. More specifically, these classes correspond to the recordings of normal sinus rhythm, atrial fibrillation (AF), alternative rhythm, or others (too noisy to be classified). The recordings are sampled at 300 Hz.
Furthermore, the dataset is imbalanced, with much fewer samples from the atrial fibrillation and noisy classes out of all four. To preprocess the dataset, we use the code from the CLOCS paper, which applied a fixed-length window of 1,500 observations to divide long recordings into short samples of 5 seconds that are physiologically meaningful. Because of the imbalanced dataset, we report AUROC and AUPRC (insensitive to label distribution) in our results. The raw dataset (\url{https://physionet.org/content/challenge-2017/1.0.0/}) is distributed under the Open Data Commons Attribution License v1.0.

\textbf{\textsc{Emg}}~\cite{goldberger_physiobank_2000}. Electromyograms (EMGs) measure muscle responses as electrical activity in response to neural stimulation, and they can be used to diagnose certain muscular dystrophies and neuropathies. The dataset consists of single-channel EMG recording from the tibialis anterior muscle of three healthy volunteers suffering from neuropathy and myopathy. The recordings are sampled at a frequency of 4K Hz. Each patient (\ie, disorder) is a classification category.
Then the recordings are split into time-series samples using a fixed-length window of 1,500 observations. The raw dataset (\url{https://physionet.org/content/emgdb/1.0.0/}) is distributed under the Open Data Commons Attribution License v1.0.


\subsection{Varying lengths and dimensionality of time series in pre-training vs.~target datasets}

We use established strategies to address the challenging question of how to process time series with the variable number of dimensions (channels) and measurements (note that this study focuses on designing a pre-training model that can transfer knowledge across disparate time-series datasets). When the dimensionality or length of time series is different between pre-training and target datasets, we view the pre-training dataset as an anchor according to which we adjust (pre-process) the target dataset. For length adjustments, we use zero padding to increase the number of observations or use downsampling to reduce the number of observations. For dimensionality adjustments, our model applies to time series at the level of a single lead, enabling the ability to generalize to multivariate time series when we make different channels share the model (\ie, time encoder, frequency encoder, and two projectors) parameters. In this way, we have a $\bm{z}^{\textsc{tune}}_i$ for each channel of the time series (Sec.~\ref{sec:method}) and concatenate the learned representations across all channels to form the final representation of the input time series sample.

\begin{table}[]
\centering
\def\arraystretch{1.10}
\caption{Performance on one-to-one setting (scenario 1): pre-training on \textbf{\textsc{SleepEeg}} and fine-tuning on \textbf{\textsc{Epilepsy}}. 
}
\label{tab:sleep2epilepsy}
\resizebox{\textwidth}{!}{%
\begin{tabular}{Hllrrrrr}
\toprule
 & \textbf{Models} & \textbf{Accuracy} & \multicolumn{1}{l}{\textbf{Precision}} & \multicolumn{1}{l}{\textbf{Recall}} & \multicolumn{1}{l}{\textbf{F1 score}} & \multicolumn{1}{l}{\textbf{AUROC}} & \multicolumn{1}{l}{\textbf{AUPRC}} \\ \midrule
\multirow{9}{*}{\begin{tabular}[c]{@{}l@{}}\textbf{\textsc{SleepEeg}}\\ $\rightarrow$\\ \\ \textbf{\textsc{Epilepsy}}\end{tabular}} & \textbf{Non-DL (KNN)} & 0.8525\std{0.0000} & \multicolumn{1}{l}{0.8639\std{0.0000}} & \multicolumn{1}{l}{0.6431\std{0.0000}} & \multicolumn{1}{l}{0.6791\std{0.0000}} & \multicolumn{1}{l}{0.6434\std{0.0000}} & \multicolumn{1}{l}{0.6279\std{0.0000}} \\
 & \textbf{Random Init.} & 0.8983\std{0.0656} & \multicolumn{1}{l}{0.9213\std{0.1369}} & \multicolumn{1}{l}{0.7447\std{0.1135}} & \multicolumn{1}{l}{0.7959\std{0.1208}} & \multicolumn{1}{l}{0.8578\std{0.2153}} & \multicolumn{1}{l}{0.6489\std{0.1926}} \\ \cmidrule{2-8}
 & \textbf{TS-SD} & 0.8952\std{0.0522} & 0.8018\std{0.2244} & 0.7647\std{0.1485} & 0.7767\std{0.1855} & 0.7677\std{0.2452} & 0.7940\std{0.1825} \\
 & \textbf{TS2vec} & 0.9395\std{0.0044} & 0.9059\std{0.0116} & 0.9039\std{0.0118} & 0.9045\std{0.0067} & 0.9587\std{0.0086} & 0.9430\std{0.0103} \\
 & \textbf{CLOCS} & \textbf{0.9507\std{0.0027}} & 0.9301\std{0.0067} & \textbf{0.9127\std{0.0165}} & \textbf{0.9206\std{0.0066}} & 0.9803\std{0.0023} & 0.9609\std{0.0116} \\
 & \textbf{Mixing-up} & 0.8021\std{0.0000} & 0.4011\std{0.0000} & 0.5000\std{0.0000} & 0.4451\std{0.0000} & 0.9743\std{0.0081} & 0.9618\std{0.0104} \\
 & \textbf{TS-TCC} & 0.9253\std{0.0098} & 0.9451\std{0.0049} & 0.8181\std{0.0257} & 0.8633\std{0.0215} & 0.9842\std{0.0034} & 0.9744\std{0.0043} \\
 & \textbf{SimCLR} & 0.9071\std{0.0344} & 0.9221\std{0.0166} & 0.7864\std{0.1071} & 0.8178\std{0.0998} & 0.9045\std{0.0539} & 0.9128\std{0.0205} \\
 & \textbf{TNC}  & 0.6701\std{0.0071} & 0.5336\std{0.2742} &0.5011\std{0.0016} &0.4024\std{0.0033} &0.5853 	\std{ 0.0487} &0.6276\std{0.0438} \\
&\textbf{CPC}  & 0.662\std{ 0.0110}&0.4340\std{0.2215}& 0.5028\std{0.0064} &0.406 	\std{ 0.0134}&0.5722\std{0.0748} & 0.6145\std{0.0575}\\
 & \textbf{TF-C (Ours)} & 0.9495\std{0.0249} & \textbf{0.9456\std{0.0108}} & 0.8908\std{0.0216} & 0.9149\std{0.0534} & \textbf{0.9811\std{0.0237}} & \textbf{0.9703\std{0.0199}} \\
\bottomrule
\end{tabular}
}
\end{table}

\begin{table}[]
\centering
\def\arraystretch{1.10}
\caption{Performance on one-to-one setting (scenario 2): pre-training on \textbf{\textsc{Fd-a}} and fine-tuning on \textbf{\textsc{Fd-b}}. 
}
\label{tab:fda2b}
\resizebox{\textwidth}{!}{%
\begin{tabular}{Hllrrrrr}
\toprule
 & \textbf{Models} & \textbf{Accuracy} & \multicolumn{1}{l}{\textbf{Precision}} & \multicolumn{1}{l}{\textbf{Recall}} & \multicolumn{1}{l}{\textbf{F1 score}} & \multicolumn{1}{l}{\textbf{AUROC}} & \multicolumn{1}{l}{\textbf{AUPRC}} \\ \midrule
\multirow{9}{*}{\begin{tabular}[c]{@{}l@{}}\textbf{\textsc{Fd-a}}\\ $\rightarrow$\\ \\ \textbf{\textsc{Fd-b}}\end{tabular}} 
& \textbf{Non-DL (KNN)} & 0.4473\std{0.0000} & \multicolumn{1}{l}{0.2846\std{0.0000}} & \multicolumn{1}{l}{0.3275\std{0.0000}} & \multicolumn{1}{l}{0.2284\std{0.0000}} & \multicolumn{1}{l}{0.4946\std{0.0000}} & \multicolumn{1}{l}{0.3307\std{0.0000}} \\
 & \textbf{Random Init.} & 0.4736\std{0.1085} & \multicolumn{1}{l}{0.4829\std{0.1235}} & \multicolumn{1}{l}{0.5235\std{0.0956}} & \multicolumn{1}{l}{0.4911\std{0.1123}} & \multicolumn{1}{l}{0.7864\std{0.2659}} & \multicolumn{1}{l}{0.7528\std{0.1233}} \\
 & \textbf{TS-SD} & 0.5465\std{0.0417} & 0.5795\std{0.0608} & 0.6613\std{0.0324} & 0.5567\std{0.0478} & 0.7148\std{0.0014} & 0.6370\std{0.0231} \\
 & \textbf{TS2vec} & 0.6494\std{0.0379} & 0.6614\std{0.0599} & 0.7260\std{0.0377} & 0.6797\std{0.0553} & 0.8089\std{0.0254} & 0.7138\std{0.0593} \\
 & \textbf{CLOCS} & 0.7118\std{0.0361} & 0.7607\std{0.0830} & 0.7877\std{0.0294} & 0.7638\std{0.0696} & 0.8161\std{0.0146} & 0.8133\std{0.0295} \\
 & \textbf{Mixing-up} & 0.7821\std{0.0110} & 0.8887\std{0.0039} & 0.8404\std{0.0081} & 0.8307\std{0.0104} & 0.9339\std{0.0003} & 0.9348\std{0.0010} \\
 & \textbf{TS-TCC} & 0.8497\std{0.0114} & 0.8906\std{0.0087} & \textbf{0.8899\std{0.0083}} & 0.8898\std{0.0083} & 0.9145\std{0.0138} & 0.8957\std{0.0198} \\
 & \textbf{SimCLR} & 0.5439\std{0.0562} & 0.6372\std{0.0568} & 0.6128\std{0.1344} & 0.5799\std{0.1286} & 0.7383\std{0.0622} & 0.6548\std{0.1152} \\
 & \textbf{TF-C (Ours)} & \textbf{0.8934\std{0.0379}} & \textbf{0.9209\std{0.0234}} & 0.8537\std{0.0486} & \textbf{0.9162\std{0.0826}} & \textbf{0.9435\std{0.0259}} & \textbf{0.9527\std{0.0134}} \\
\bottomrule
\end{tabular}
}
\end{table}

\begin{table}[]
\centering
\def\arraystretch{1.10}
\caption{Performance on one-to-one setting (scenario 4): pre-training on \textbf{\textsc{Ecg}}  and fine-tuning on \textbf{\textsc{Emg}}. 
}
\label{app:ecg2emg}
\resizebox{\textwidth}{!}{%
\begin{tabular}{Hllrrrrr}
\toprule
 & \textbf{Models} & \textbf{Accuracy} & \multicolumn{1}{l}{\textbf{Precision}} & \multicolumn{1}{l}{\textbf{Recall}} & \multicolumn{1}{l}{\textbf{F1 score}} & \multicolumn{1}{l}{\textbf{AUROC}} & \multicolumn{1}{l}{\textbf{AUPRC}} \\ \midrule
\multirow{9}{*}{\begin{tabular}[c]{@{}l@{}}\textbf{\textsc{Ecg}} \\ $\rightarrow$\\ \\ \textbf{\textsc{Emg}}\end{tabular}} 
& \textbf{Non-DL (KNN)} & 0.4390\std{0.0000} & \multicolumn{1}{l}{0.3771\std{0.0000}} & \multicolumn{1}{l}{0.5143\std{0.0000}} & \multicolumn{1}{l}{0.3979\std{0.0000}} & \multicolumn{1}{l}{0.6025\std{0.0000}} & \multicolumn{1}{l}{0.4083\std{0.0000}} \\
 & \textbf{Random Init.} & 0.878\std{0.1259} & \multicolumn{1}{l}{0.5909\std{0.1135}} & \multicolumn{1}{l}{0.6667\std{0.1534}} & \multicolumn{1}{l}{0.6238\std{0.2315}} & \multicolumn{1}{l}{0.9109\std{0.1264}} & \multicolumn{1}{l}{0.7771\std{0.1359}} \\
 & \textbf{TS-SD} & 0.4606\std{0.0000} & 0.1544\std{0.0000} & 0.3333\std{0.0000} & 0.2111\std{0.0000} & 0.5031\std{0.0219} & 0.3805\std{0.0165} \\
 & \textbf{TS2vec} & 0.9704\std{0.0109} & 0.9666\std{0.0186} & 0.9751\std{0.0082} & \textbf{0.9746\std{0.0141}} & 0.9948\std{0.0070} & \textbf{0.9780\std{0.0305}} \\
 & \textbf{CLOCS} & 0.8829\std{0.0499} & 0.8492\std{0.0620} & 0.8134\std{0.1262} & 0.8037\std{0.1080} & 0.9385\std{0.0369} & 0.8178\std{0.0945} \\
 & \textbf{Mixing-up} & 0.9121\std{0.0872} & 0.8007\std{0.2158} & 0.8518\std{0.1776} & 0.8215\std{0.2010} & \textbf{0.9999\std{0.0000}} & 0.9999\std{0.0000} \\
 & \textbf{TS-TCC} & 0.9590\std{0.0135} & \textbf{0.9684\std{0.0140}} & 0.8994\std{0.0304} & 0.9244\std{0.0247} & 0.9800\std{0.0192} & 0.9663\std{0.0348} \\
 & \textbf{SimCLR} & 0.8878\std{0.0218} & 0.8209\std{0.0307} & 0.8533\std{0.0433} & 0.8225\std{0.0339} & 0.9565\std{0.0153} & 0.8337\std{0.0163} \\
 & \textbf{TF-C (Ours)} & \textbf{0.9756\std{0.0071}} & 0.9444\std{0.0029} & \textbf{0.9803\std{0.0000}} & 0.9596\std{0.0003} & 0.9801\std{0.0012} & 0.8867\std{0.0122} \\
\bottomrule
\end{tabular}
}
\end{table}

\section{TF-C and baseline architectures and implementation details}
\label{app:baseline_implementation}
We compare our TF-C model against eight state-of-the-art baselines and two additional methods (Sec.~\ref{sec:experiments}): (1) directly fit a KNN (K=2) classifier with the target (fine-tuning) dataset; (2) randomly initialize a model (the model structure and experimental settings are the same with our TF-C model), ignore pre-training, and directly train from scratch with the target dataset.

We implement the baselines follow the corresponding papers including TS-SD~\cite{shi_self-supervised_2021}, TS2vec~\cite{yue_ts2vec_2022}, CLOCS~\cite{kiyasseh_clocs_nodate}, Mixing-up~\cite{wickstrom_mixing_2022}, TS-TCC~\cite{eldele_time-series_2021}, SimCLR~\cite{tang_exploring_2021}, TNC~\cite{tonekaboni_unsupervised_2021}, and CPC~\cite{oord_representation_2018}. 
We use default settings for hyper-parameters as reported in the original works unless noted below. All pre-training and fine-tuning with baselines are done with a single Tesla V100 GPU with 32 Gb of allocated memory provided by Harvard Medical School's O2 High Performance Computing platform. 

\textbf{TF-C (our model)} Our time-based contrastive encoder $F_{\textsc{T}}$ adopts 1-D ResNet backbone similar to SimCLR. Specifically, after hyper-parameter tuning, the encoder consists of three layers convolutional blocks: the kernel sizes in all layers are 8; the strides are 8, 1, and 1, respectively; the the depths are 32, 64, and 128, respectively. We use max pooling after each convolutional layer and all the pooling kernel sizes and strides are set as 2. The cross-space projector $R_{\textsc{T}}$ contains two fully-connected layers with hidden dimensions 256 and 128, respectively. 
In the transformation from time space to frequency space, we use the full spectrum (symmetrical) thus $\bm{x}^{\textsc{T}}$ and $\bm{x}^{\textsc{F}}$ have the same dimension. For simplify, the frequency encoder $F_{\textsc{F}}$ and projector $R_{\textsc{F}}$ have same structure (but different parameters) with their counterparts in time space. Our preliminary experiments show that the change of model structure has relatively small change on performance. For example, replacing $F_{\textsc{T}}$ by 2-layer LSTM or 3-layer Transformer only cause a slight drop of F1 score ($\pm$ 1.2\%; one-to-one setting; Scenario 1). In pre-training, we use Adam optimizer with learning rate of 0.0003 and 2-norm penalty coefficient of 0.0005. We use batch size of 64 and training epoch of 40. We record the best performance in terms of F1 score and save the parameters in the corresponding epoch as $\Theta$. In fine-tuning, we initialize the model parameters with $\Theta$ and optimize the model on fine-tuning set. The hyper-parameters are the same as in pre-training. For the classification task in fine-tuning stage, we adopt a 2-layer fully-connected layer as a classifier. The hidden dimensions are 64 and the number classes in target dataset, respectively. We measure the classification loss via the cross entropy function and the loss is jointly optimized with the fine-tuning loss (\ie, $\mathcal{L}_{\textsc{TF-C}}$).

\textbf{TS-SD}~\cite{shi_self-supervised_2021} uses a modified attention mechanism to encode latent features and use denoising or DTW-similarity prediction as pretext-tasks.The encoder network is based on self-attention mechanism~\cite{zerveas2021transformer} but instead of the linear layer that produce the K, Q, V matrices from the input sequence, TS-SD uses a single convolutional layer to replace the linear layer. With multihead attention, TS-SD sets a different kernel size for each head to attempt capture short- and long-range temporal patterns. Although we found that 12 heads was relatively redundant and may be replaced by fewer heads, we still use 12 heads to keep the same with the original work in \cite{shi_self-supervised_2021}. For self-supervised pre-training, we adopt the denoising pretext task that attempts to minimize the mean square distance between the original input time series and the output of the encoder, for which the input is an augmented time series sample with noise added to a subseries. To cope with small numerical values, we use a learning rate of 3e-7 in pre-training and 3e-4 during fine-tuning. 
While training on pre-training dataset for the four scenarios,
the batch sizes varied from 4 to 128 and were manually chosen to achieve reasonable performance for each scenario. During fine-tuning, we use \# epochs ranging from 20 to 80 and a batch size of 16. Although the source code of TS-SD is not released, we implement the TS-SD model to our best understanding and tune the model to achieve the best performance as a strong baseline. We public the implemented TS-SD in our online repository.   

\textbf{TS2vec}~\cite{yue_ts2vec_2022} introduces the notion of contextual consistency and uses a hierarchical loss function to capture long-range structure in time series. TS2vec is a powerful representative learning method and have a specially-designed architecture. The encoder network consists of three components. First, the input time series is augmented by selecting overlapping subseries. They are projected into a higher dimensional latent space. Then, latent vectors for input time series are masked at randomly chosen positions. Finally, a dilated CNN with residual blocks produce the contextual representations. To compute the loss, the representations are gradually pooled along the time dimension and at each step a loss function based on contextual consistency is applied. For the baseline experiment, we found that the original 10 layers of ResNet blocks is redundant, and we reduce residual blocks in the encoder from 10-layer to 2-layer without compromising model performances. To make the model comparable with other baselines, we also restrict 
the hidden dimension to 1 and use a batch size of 64, except in the second scenario we reduce it to 16 given the length of samples in the \textbf{\textsc{Fd-a}}/\textbf{\textsc{Fd-b}} datasets.


\textbf{CLOCS}~\cite{kiyasseh_clocs_nodate} assumes that samples from different subjects are negative pairs in contrastive learning and applies it to classify ECG signals.
CLOCS makes the basic assumption that time series samples from different patients (the application scenario is in learning cardiac signals) should have dissimilar representations. Based on that, they devised three slightly different architectures - CMSC, CMLC, and CMSMLC, where ECG recordings from non-overlapping temporal segments, different ECG leads, and both of these are treated as positive pairs in case if they come from the same patients. For this model, the encoder consists of alternating convolutional and pooling layers. So given the different input lengths, the hyper-parameters for these layers have to be modified accordingly. For example, we use kernel sizes of $7,7,7$, stride of $3,3,3$, pooling layer size of $2$ for the last scenario of \textbf{\textsc{Ecg}}  to \textbf{\textsc{Emg}}  but increased stride to $4$ and increased representation size for a time series sample from $64$ to $128$ in case of \textbf{\textsc{Fd-a}} to \textbf{\textsc{Fd-b}} due to much longer time series samples. 

\textbf{Mixing-up}~\cite{wickstrom_mixing_2022} proposes new mixing-up augmentation and pretext tasks that aim to correctly predict the mixing proportion of two time series samples.
In Mixing-up, the augmentation is chosen as the convex combination of two randomly drawn time series from the dataset, where the mixing parameter is random drawn from a beta distribution. The contrastive loss is then computed between the two inputs and the augmented time series. The loss is a minor modification of NT-Xent loss and is designed to encourage the correct prediction of the amount of mixing. We use the same beta distribution as reported in the original Mixing-up model. 

\textbf{TS-TCC}~\cite{eldele_time-series_2021} leverages contextual information with a transformer-based autoregressive model and ensures transferability by using both strong and weak augmentations.
TS-TCC proposed a challenging pretext task. An input time series sample is first augmented by adding noise, scaling, and permuting time series. The views are then passed through an encoder consisting of three convolutional layers before processed by the temporal contrasting module. During temporal contrasting, for each view, a transformer architecture is used to learn a contextual representation. The learned representation is then used to predict latent observation of the other augmented view at a future time. The contextual representations are then projected and maximized similarity using NT-Xent loss. For this baseline, we mostly adopted the hyper-parameters presented in the original paper. We use a learning rate of 3e-4 throughout, and a batch size of 128 during pre-training and 16 during fine-tuning. 

\textbf{SimCLR}~\cite{tang_exploring_2021} is a state-of-the-art model in self-supervised representation learning of images. It utilizes deep learning architectures to generate augmentation-based embeddings and optimize the model parameters by minimizing NT-Xent loss in the embedding space. \cite{tang_exploring_2021} applies the original SimCLR model for time series data. In this work, we compare with the modified SimCLR as in \cite{tang_exploring_2021}.
The SimCLR contrastive learning framework consists of four major components. Although initially proposed for image data, it is readily adapted to time series, as shown in \cite{tang_exploring_2021}. An input time series sample is first stochastically augmented into two related views. Then a base encoder extracts representation vectors. ResNet is used as encoder backbone for simplicity. Then the projection head transforms representations into a latent space where the NT-Xent loss is applied. For time series, SimCLR investigated different augmentations, including adding noise, scaling, rotation, negation, flipping in time, permutation of subseries, time warping, and channel shuffling. All the unmentioned hyper-parameters are kept the same with the original model.

\textbf{TNC}~\cite{tonekaboni_unsupervised_2021} focuses on learning representations that encode the underlying state of a non-stationary time series. This is achieved by ensuring that the distribution of observations in the latent space is different from the distribution of temporally separated observations. For each time point, the TNC model calculates a neighborhood size using the Augmented Dickey–Fuller test. The encoded representations of two different windows will be passed to a discriminator to predict the probability that they belong to the same temporal neighborhood. 
We implement the TFC model following the public GitHub repository provided by the authors. We observed that the ADF tests are extremely slow, so TNC is only used as a baseline in scenario 1 in one-to-one setting (classification task). 

\textbf{CPC}~\cite{oord_representation_2018} aims to learn the representations that encode the global information shared across different segments of a time series signal by the strategy of predictive coding. Instead of modelling the conditional distribution at future time points for prediction, the authors choose to encode the context and observation into a vector representation in a way that maximally preserves their mutual information. For the model architecture, a nonlinear encoder first maps the sequence of observations to latent representations. Then an autoregressive model summarizes past latent representations to produce a context representation at every time point. The two components are trained to jointly optimize a loss function which is inspired by noise contrastive estimation.
We used the implementation provided by the authors of the TNC paper \cite{tonekaboni_unsupervised_2021} and kept their default choices of parameters whenever reasonable.

The TS-SD is the only method, to our knowledge, that explicitly aims at time series pre-training but the underlying assumption that can support knowledge transfer across different time series datasets is not clearly provided. The TS2vec, CLOCS, Mixing-up, TS-TCC, TNC, and CPC are designed for representation learning (instead of pre-training) but involved transfer learning setting in their experiments. We modify them to perform pre-training task and make them comparable to our model. The SimCLR is a widely use baseline in computer vision and simply adopted to deal with time series by \cite{tang_exploring_2021}. We compare our model with SimCLR to further show our superiority. 


\section{Additional results on one-to-one evaluation}
\label{app:one_to_one_details}
In one-to-one pre-training evaluation, we consider four different scenarios with paired pre-training and fine-tuning datasets from different fields (Sec.~\ref{sub:one_to_one}). In each scenario, the pre-training and fine-tuning datasets share same semantic meanings (\eg, both EEG signals; Scenario 1) or similar semantic meanings (\eg, ECG and EMG are both physiological signals; Scenario 4).
For example, in the first scenario, \textbf{\textsc{SleepEeg}} and \textbf{\textsc{Epilepsy}} both involve EEG measurements but the recordings are taken under different physical conditions.

For each scenario, we first train each baseline model using the pre-training dataset and record the converged parameters. The parameters are then taken to initialize the model on the fine-tuning phase which uses the target dataset. In fine-tuning, we allowed all parameters in the model to be optimized (no freezing of earlier layers; full fine-tuning). Our preliminary results show that the performance of full fine-tuning is slightly better than partial fine-tuning by 5.4\% in F1 score (one-to-one setting; Scenario 1).
After fine-tuning, we evaluate the final model on the test set and report metrics including accuracy, precision (macro-averaged), recall, F1-score, AUROC (one-versus-rest), and AUPRC. The evaluation results of four scenarios are shown in Table~\ref{tab:har2Gesture} (main paper) and Tables~\ref{tab:sleep2epilepsy}-\ref{app:ecg2emg}.

\section{Results on downstream clustering and anomaly detection tasks}
\label{app:clustering_anomaly}
Beyond the common downstream classification tasks, we conducted extensive experiments to examine the proposed TF-C in diverse downstream tasks. The comparison results with state-of-the-art baselines are shown in Table~\ref{tab:clustering} for clustering and in Table~\ref{tab:anomaly} for anomaly detection. The experimental setups and results analysis are shown in Section~\ref{sec:experiments}.

\begin{table}[]
\centering
\scriptsize
\def\arraystretch{1.0}
\caption{\textbf{Performance on downstream clustering.} 
Pre-training is performed on \textsc{SleepEeg} dataset, followed by an independent fine-tuning on \textsc{Epilepsy}. We compare with five baselines including two non-transfer-based baselines (Random Init. and Non-DL), the best performing baseline in the context of classification task (i.e., TS-TCC), and two new models (TNC and CPC).
}
\label{tab:clustering}
\begin{tabular}{llll}
\toprule
\textbf{Method}       & \textbf{Silhouette} & \textbf{ARI}      & \textbf{NMI}      \\ \midrule
\textbf{Non-DL(KNN)}  & 0.1208\std{0.0271}    & 0.0549\std{ 0.0059} & 0.0096\std{ 0.0014}  \\
\textbf{Random Init.} & 0.3497\std{ 0.0509}   & 0.3216\std{0.0136}  & 0.2408\std{ 0.0287}  \\ \midrule
\textbf{TNC}          & 0.2353\std{ 0.0018}   & 0.0211\std{ 0.0061}  & 0.0082\std{ 0.0018}  \\
\textbf{CPC}          & 0.2223\std{0.0011}    & 0.0153\std{ 0.0196} & 0.0063\std{ 0.0055} \\
\textbf{TS-TCC}       & 0.5154\std{ 0.0458}   & 0.6307\std{ 0.0325} & 0.5178\std{ 0.0283} \\
\textbf{TF-C (Ours)}  & 0.5439\std{ 0.0417}   & 0.6583\std{0.0259}  & 0.5567\std{0.0172}  \\
\bottomrule
\end{tabular}
\end{table}

\begin{table}[]
\centering
\scriptsize
\def\arraystretch{1.0}
\caption{\textbf{Performance on downstream anomaly detection.} 
We pre-train the model on \textsc{Fd-a} and fine-tune on an anomaly detection subset of \textsc{Fd-b}. The subset is highly imbalance (90\% normal samples and 10\% abnormal samples). We evaluate the performance with precision, recall, F1 score and AUROC. 
}
\label{tab:anomaly}
\begin{tabular}{lllll}
\toprule
\textbf{Models}       & \textbf{Precision} & \textbf{Recall}  & \textbf{F1 score} & \textbf{AUROC}  \\ \midrule
\textbf{Non-DL(KNN)}  & 0.4785\std{0.0356}    & 0.6159\std{0.0585}  & 0.5061\std{0.0278}    & 0.7653\std{0.0153} \\
\textbf{Random Init.} & 0.8219\std{0.05319}   & 0.7131\std{0.02773} & 0.7639\std{0.0094}    & 0.8174\std{0.0218} \\ \midrule
\textbf{TNC}          & 0.8354\std{0.0484}    & 0.7882\std{0.0157}  & 0.7957\std{0.0165}    & 0.8231\std{0.0379} \\
\textbf{CPC}          & 0.5967\std{0.0776}    & 0.4896\std{0.0866}  & 0.5238\std{0.0792}    & 0.7859\std{0.0356} \\
\textbf{TS-TCC}       & 0.6219\std{0.0183}    & 0.4431\std{0.0658}  & 0.4759\std{0.0562}    & 0.7966\std{0.0441} \\
\textbf{TF-C (Ours)}  & 0.8526\std{0.0367}   & 0.7823\std{0.0299 } & 0.8312\std{0.0186}    & 0.8598\std{0.0283} \\
\bottomrule
\end{tabular}
\end{table}

\section{Further results on ablation study}
\label{app:ablation}
We conduct ablation studies (Sec.~\ref{sec:experiments}) to evaluate the importance of every component in the developed TF-C model. For the example of one-to-one setting (\textbf{\textsc{SleepEeg}} $\rightarrow$ \textbf{\textsc{Gesture}}) when pre-training the model using \textbf{\textsc{SleepEeg}} dataset and fine-tuning on \textbf{\textsc{Gesture}} dataset, ablation study results are shown in Table~\ref{tab:ablation}. Through comparison, we observe that the full TF-C model achieves the highest performance in every evaluation metrics, indicating every component, especially the novel consistency loss $\mathcal{L}_{\textsc{C}}$), contributes to the model's performance.

\begin{table}[t]
\centering
\def\arraystretch{1.10}
\caption{Ablation study (\textbf{\textsc{SleepEeg}} $\rightarrow$ \textbf{\textsc{Epilepsy}}).
If we remove $\mathcal{L}_{\textsc{T}}$, we cannot calculate $\mathcal{L}_{\textsc{C}}$, thus we remove both $\mathcal{L}_{\textsc{T}}$ and $\mathcal{L}_{\textsc{C}}$ (instead of only remove $\mathcal{L}_{\textsc{T}}$). The same for $\mathcal{L}_{\textsc{F}}$.
"W/o $\mathcal{L}_{\textsc{C}}$ and $\mathcal{L}_{\textsc{T}}$" means removing the time encoder (Sec.~\ref{sub:time_contrastive}) and the consistency module (Sec.~\ref{sub:TF-C}), using $\bm{z}^{\textsc{F}}$ as final embedding for downstream tasks in fine-tuning. 
"W/o $\mathcal{L}_{\textsc{C}}$ and $\mathcal{L}_{\textsc{F}}$" means removing the frequency encoder (Sec.~\ref{sub:frequency_contrastive}) and the consistency module, using $\bm{z}^{\textsc{T}}$ as final embedding for downstream tasks in fine-tuning. 
W/o $\mathcal{L}_{\textsc{C}}$ means removing the consistency module, using $[\bm{z}^{\textsc{T}};\bm{z}^{\textsc{F}}]$ as the final embedding for downstream tasks in fine-tuning.
"Replace $\mathcal{L}_{\textsc{C}}$ with $\mathcal{L}_{\textsc{FF-C}}$" refers to replacing Eq.~\ref{eq:loss_c} by the distance $d(\bm{z}^{\textsc{f}}_i, \widetilde{\bm{z}}^{\textsc{f}}_i, \mathcal{D}^{\textrm{pret}})$ between two frequency-based embeddings. 
"Replace $\mathcal{L}_{\textsc{C}}$ with $\mathcal{L}_{\textsc{TT-C}}$" refers to replacing Eq.~\ref{eq:loss_c} by the distance $d(\bm{z}^{\textsc{t}}_i, \widetilde{\bm{z}}^{\textsc{t}}_i, \mathcal{D}^{\textrm{pret}})$ between two time-based embeddings.
}
\label{tab:ablation}
\resizebox{\textwidth}{!}{%
\begin{tabular}{lllllll}
\toprule
& \textbf{Accuracy}       & \textbf{Precision}      & \textbf{Recall}         & \textbf{F1 score}       & \textbf{AUROC}          & \textbf{AUPRC}          \\ \midrule
W/o $\mathcal{L}_{\textsc{C}}$ and $\mathcal{L}_{\textsc{T}}$      & 0.7159$\pm$0.0128          & 0.7211$\pm$0.0428          & 0.7246$\pm$0.0428          & 0.7239$\pm$0.0429          & 0.8597$\pm$0.0236          & 0.7655$\pm$0.0386          \\
W/o $\mathcal{L}_{\textsc{C}}$ and $\mathcal{L}_{\textsc{F}}$      & 0.7327$\pm$0.0328          & 0.7246$\pm$0.0311          & 0.7339$\pm$0.0307          & 0.7317$\pm$0.0356          & 0.8991$\pm$0.0279          & 0.7236$\pm$0.0278          \\
W/o $\mathcal{L}_{\textsc{C}}$              & 0.7428$\pm$0.0297          & 0.7289$\pm$0.0278          & 0.7451$\pm$0.0263          & 0.7377$\pm$0.0308          & 0.9125$\pm$0.0167          & 0.7706$\pm$0.0135          \\
Replace $\mathcal{L}_{\textsc{C}}$ with $\mathcal{L}_{\textsc{FF-C}}$ & 0.7259$\pm$0.0072          & 0.7319$\pm$0.0256          & 0.7338$\pm$0.0133          & 0.7341$\pm$0.0194          & 0.9015$\pm$0.0135          & 0.7529$\pm$0.0096          \\
Replace $\mathcal{L}_{\textsc{C}}$ with $\mathcal{L}_{\textsc{TT-C}}$ & 0.7124$\pm$0.0091          & 0.7256$\pm$0.0169          & 0.7231$\pm$0.0197          & 0.7296$\pm$0.0209          & 0.8726$\pm$0.0098          & 0.7627$\pm$0.0107          \\
Full Model (TF-C)      & \textbf{0.7642$\pm$0.0196} & \textbf{0.7731$\pm$0.0355} & \textbf{0.7429$\pm$0.0268} & \textbf{0.7572$\pm$0.0311} & \textbf{0.9238$\pm$0.0159} & \textbf{0.7961$\pm$0.0109} \\ \bottomrule
\end{tabular}
}
\end{table}

\begin{table}[]
\centering
\def\arraystretch{0.96}
\caption{\textbf{One-to-many pre-training evaluation.} Pre-training is performed on \textsc{SleepEeg} dataset, followed by an independent fine-tuning on \textsc{Epilepsy}, \textsc{Fd-b}, \textsc{Gesture}, and \textsc{Emg}.}
\label{tab:one_to_many}
\resizebox{\textwidth}{!}{%
\begin{tabular}{cllrrrrr}
\toprule \toprule
 \textbf{Scenarios} & \textbf{Models} & \textbf{Accuracy} & \multicolumn{1}{l}{\textbf{Precision}} & \multicolumn{1}{l}{\textbf{Recall}} & \multicolumn{1}{l}{\textbf{F1 score}} & \multicolumn{1}{l}{\textbf{AUROC}} & \multicolumn{1}{l}{\textbf{AUPRC}} \\ \midrule
\multirow{9}{*}{\begin{tabular}[c]{@{}l@{}}\;\textsc{SleepEeg}\\ $\quad\;\;\; \downarrow$\\\;\textsc{Epilepsy}\end{tabular}} & \textbf{Non-DL (KNN)} & 0.8525\std{0.0000} & \multicolumn{1}{l}{0.8639\std{0.0000}} & \multicolumn{1}{l}{0.6431\std{0.0000}} & \multicolumn{1}{l}{0.6791\std{0.0000}} & \multicolumn{1}{l}{0.6434\std{0.0000}} & \multicolumn{1}{l}{0.6279\std{0.0000}} \\
 & \textbf{Random Init.} & 0.8983\std{0.0656} & \multicolumn{1}{l}{0.9213\std{0.1369}} & \multicolumn{1}{l}{0.7447\std{0.1135}} & \multicolumn{1}{l}{0.7959\std{0.1208}} & \multicolumn{1}{l}{0.8578\std{0.2153}} & \multicolumn{1}{l}{0.6489\std{0.1926}} \\ \cmidrule{2-8}
 & \textbf{TS-SD} & 0.8952\std{0.0522} & 0.8018\std{0.2244} & 0.7647\std{0.1485} & 0.7767\std{0.1855} & 0.7677\std{0.2452} & 0.7940\std{0.1825} \\
 & \textbf{TS2vec} & 0.9395\std{0.0044} & 0.9059\std{0.0116} & 0.9039\std{0.0118} & 0.9045\std{0.0067} & 0.9587\std{0.0086} & 0.9430\std{0.0103} \\
 & \textbf{CLOCS} & \textbf{0.9507\std{0.0027}} & 0.9301\std{0.0067} & \textbf{0.9127\std{0.0165}} & \textbf{0.9206\std{0.0066}} & 0.9803\std{0.0023} & 0.9609\std{0.0116} \\
 & \textbf{Mixing-up} & 0.8021\std{0.0000} & 0.4011\std{0.0000} & 0.5000\std{0.0000} & 0.4451\std{0.0000} & 0.9743\std{0.0081} & 0.9618\std{0.0104} \\
 & \textbf{TS-TCC} & 0.9253\std{0.0098} & 0.9451\std{0.0049} & 0.8181\std{0.0257} & 0.8633\std{0.0215} & 0.9842\std{0.0034} & 0.9744\std{0.0043} \\
 & \textbf{SimCLR} & 0.9071\std{0.0344} & 0.9221\std{0.0166} & 0.7864\std{0.1071} & 0.8178\std{0.0998} & 0.9045\std{0.0539} & 0.9128\std{0.0205} \\
 & \textbf{TF-C (Ours)} & 0.9495\std{0.0249} & \textbf{0.9456\std{0.0108}} & 0.8908\std{0.0216} & 0.9149\std{0.0534} & \textbf{0.9811\std{0.0237}} & \textbf{0.9703\std{0.0199}} \\ 
 \midrule \midrule
\multirow{9}{*}{\begin{tabular}[c]{@{}l@{}}\;\textsc{SleepEeg}\\ $\quad\;\;\; \downarrow$\\ \;\;\;\textsc{Fd-b}\end{tabular}} & \textbf{Non-DL (KNN)} & 0.4473\std{0.0000} & \multicolumn{1}{l}{0.2847\std{0.0000}} & \multicolumn{1}{l}{0.3275\std{0.0000}} & \multicolumn{1}{l}{0.2284\std{0.0000}} & \multicolumn{1}{l}{0.4946\std{0.0000}} & \multicolumn{1}{l}{0.3308\std{0.0000}} \\
 & \textbf{Random Init.} & 0.4736\std{0.0623} & \multicolumn{1}{l}{0.4829\std{0.0529}} & \multicolumn{1}{l}{0.5235\std{0.1023}} & \multicolumn{1}{l}{0.4911\std{0.0590}} & \multicolumn{1}{l}{0.7864\std{0.0349}} & \multicolumn{1}{l}{0.7528\std{0.0254}} \\ \cmidrule{2-8}
 & \textbf{TS-SD} & 0.5566\std{0.0210} & 0.5710\std{0.0535} & 0.6054\std{0.0272} & 0.5703\std{0.0328} & 0.7196\std{0.0113} & 0.5693\std{0.0532} \\
 & \textbf{TS2vec} & 0.4790\std{0.0113} & 0.4339\std{0.0092} & 0.4842\std{0.0197} & 0.4389\std{0.0107} & 0.6463\std{0.0130} & 0.4442\std{0.0162} \\
 & \textbf{CLOCS} & 0.4927\std{0.0310} & 0.4824\std{0.0316} & 0.5873\std{0.0387} & 0.4746\std{0.0485} & 0.6992\std{0.0099} & 0.5501\std{0.0365} \\
 & \textbf{Mixing-up} & 0.6789\std{0.0246} & 0.7146\std{0.0343} & \textbf{0.7613\std{0.0198}} & 0.7273\std{0.0228} & 0.8209\std{0.0035} & 0.7707\std{0.0042} \\
 & \textbf{TS-TCC} & 0.5499\std{0.0220} & 0.5279\std{0.0293} & 0.6396\std{0.0178} & 0.5418\std{0.0338} & 0.7329\std{0.0203} & 0.5824\std{0.0468} \\
 & \textbf{SimCLR} & 0.4917\std{0.0437} & 0.5446\std{0.1024} & 0.4760\std{0.0885} & 0.4224\std{0.1138} & 0.6619\std{0.0219} & 0.5009\std{0.0477} \\
 & \textbf{TF-C (Ours)} & \textbf{0.6938\std{0.0231}} & \multicolumn{1}{l}{\textbf{0.7559\std{0.0349}}} & \multicolumn{1}{l}{0.7202\std{0.0257}} & \multicolumn{1}{l}{\textbf{0.7487\std{0.0268}}} & \multicolumn{1}{l}{\textbf{0.8965\std{0.0135}}} & \multicolumn{1}{l}{\textbf{0.7871\std{0.0267}}} \\ \midrule \midrule
\multirow{9}{*}{\begin{tabular}[c]{@{}l@{}}\;\textsc{SleepEeg}\\ $\quad\;\;\;\; \downarrow$\\ \;\;\textsc{Gesture}\end{tabular}} & \textbf{Non-DL (KNN)} & 0.6833\std{0.0000} & \multicolumn{1}{l}{0.6501\std{0.0000}} & \multicolumn{1}{l}{0.6833\std{0.0000}} & \multicolumn{1}{l}{0.6443\std{0.0000}} & \multicolumn{1}{l}{0.8190\std{0.0000}} & \multicolumn{1}{l}{0.5232\std{0.0000}} \\
 & \textbf{Random Init.} & 0.4219\std{0.0629} & \multicolumn{1}{l}{0.4751\std{0.0175}} & \multicolumn{1}{l}{0.4963\std{0.0679}} & \multicolumn{1}{l}{0.4886\std{0.0459}} & \multicolumn{1}{l}{0.7129\std{0.0166}} & \multicolumn{1}{l}{0.3358\std{0.1439}} \\ \cmidrule{2-8}
 & \textbf{TS-SD} & 0.6922\std{0.0444} & 0.6698\std{0.0472} & 0.6867\std{0.0488} & 0.6656\std{0.0443} & 0.8725\std{0.0324} & 0.6185\std{0.0966} \\
 & \textbf{TS2vec} & 0.6917\std{0.0333} & 0.6545\std{0.0358} & 0.6854\std{0.0349} & 0.6570\std{0.0392} & 0.8968\std{0.0123} & 0.6989\std{0.0346} \\
 & \textbf{CLOCS} & 0.4433\std{0.0518} & 0.4237\std{0.0794} & 0.4433\std{0.0518} & 0.4014\std{0.0602} & 0.8073\std{0.0109} & 0.4460\std{0.0384} \\
 & \textbf{Mixing-up} & 0.6933\std{0.0231} & 0.6719\std{0.0232} & 0.6933\std{0.0231} & 0.6497\std{0.0306} & 0.8915\std{0.0261} & 0.7279\std{0.0558} \\
 & \textbf{TS-TCC} & 0.7188\std{0.0349} & 0.7135\std{0.0352} & 0.7167\std{0.0373} & 0.6984\std{0.0360} & 0.9099\std{0.0085} & 0.7675\std{0.0201} \\
 & \textbf{SimCLR} & 0.4804\std{0.0594} & 0.5946\std{0.1623} & 0.5411\std{0.1946} & 0.4955\std{0.1870} & 0.8131\std{0.0521} & 0.5076\std{0.1588} \\
 & \textbf{TF-C (Ours)} & \textbf{0.7642\std{0.0196}} & \multicolumn{1}{l}{\textbf{0.7731\std{0.0355}}} & \multicolumn{1}{l}{\textbf{0.7429\std{0.0268}}} & \multicolumn{1}{l}{\textbf{0.7572\std{0.0311}}} & \multicolumn{1}{l}{\textbf{0.9238\std{0.0159}}} & \multicolumn{1}{l}{\textbf{0.7961\std{0.0109}}} \\ \midrule \midrule
\multirow{9}{*}{\begin{tabular}[c]{@{}l@{}}\;\textsc{SleepEeg}\\ $\quad\;\;\; \downarrow$\\ \;\;\;\;\textsc{Emg}\end{tabular}} & \textbf{Non-DL (KNN)} & 0.4390\std{0.0000} & \multicolumn{1}{l}{0.3772\std{0.0000}} & \multicolumn{1}{l}{0.5143\std{0.0000}} & \multicolumn{1}{l}{0.3979\std{0.0000}} & \multicolumn{1}{l}{0.6025\std{0.0000}} & \multicolumn{1}{l}{0.4084\std{0.0000}} \\
 & \textbf{Random Init.} & 0.7780\std{0.0729} & \multicolumn{1}{l}{0.5909\std{0.0625}} & \multicolumn{1}{l}{0.6667\std{0.0135}} & \multicolumn{1}{l}{0.6238\std{0.0267}} & \multicolumn{1}{l}{0.9109\std{0.1239}} & \multicolumn{1}{l}{0.7771\std{0.1427}} \\ \cmidrule{2-8}
 & \textbf{TS-SD} & 0.4606\std{0.0000} & \multicolumn{1}{l}{0.1545\std{0.0000}} & 0.3333\std{0.0000} & 0.2111\std{0.0000} & 0.5005\std{0.0126} & 0.3775\std{0.0110} \\
 & \textbf{TS2vec} & 0.7854\std{0.0318} & \multicolumn{1}{l}{\textbf{0.8040\std{0.0750}}} & 0.6785\std{0.0396} & 0.6766\std{0.0501} & \textbf{0.9331\std{0.0164}} & \textbf{0.8436\std{0.0372}} \\
 & \textbf{CLOCS} & 0.6985\std{0.0323} & \multicolumn{1}{l}{0.5306\std{0.0750}} & 0.5354\std{0.0291} & 0.5139\std{0.0409} & 0.7923\std{0.0573} & 0.6484\std{0.0680} \\
 & \textbf{Mixing-up} & 0.3024\std{0.0534} & \multicolumn{1}{l}{0.1099\std{0.0126}} & 0.2583\std{0.0456} & 0.1541\std{0.0204} & 0.4506\std{0.1718} & 0.3660\std{0.1635} \\
 & \textbf{TS-TCC} & 0.7889\std{0.0192} & \multicolumn{1}{l}{0.5851\std{0.0974}} & 0.6310\std{0.0991} & 0.5904\std{0.0952} & 0.8851\std{0.0113} & 0.7939\std{0.0386} \\
 & \textbf{SimCLR} & 0.6146\std{0.0582} & \multicolumn{1}{l}{0.5361\std{0.1724}} & 0.4990\std{0.1214} & 0.4708\std{0.1486} & 0.7799\std{0.1344} & 0.6392\std{0.1596} \\
 & \textbf{TF-C (Ours)} & \textbf{0.8171\std{0.0287}} & \multicolumn{1}{l}{0.7265\std{0.0353}} & \textbf{0.8159\std{0.0289}} & \textbf{0.7683\std{0.0311}} & 0.9152\std{0.0211} & 0.8329\std{0.0137} \\ \bottomrule \bottomrule
\end{tabular}
}
\vspace{-6mm}
\end{table}

\begin{figure}
    \centering
    \includegraphics[width=\linewidth]{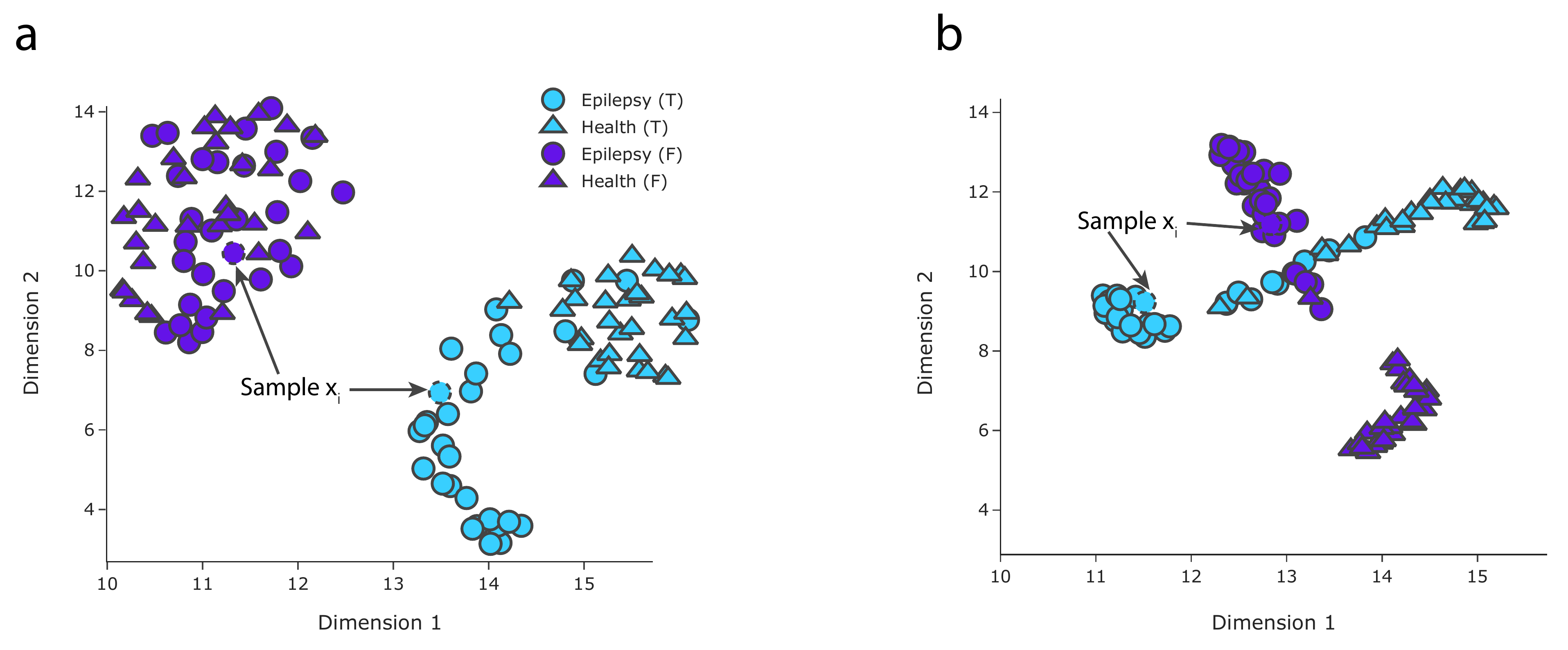}
    \caption{Visualizing time-based and frequency-based embeddings in \textbf{\textsc{Epilepsy}} (one-to-one; pre-training on \textbf{\textsc{SleepEeg}}). (a) Without consistency loss item $\mathcal{L}_{\textsc{c}}$ in both pre-training and fine-tuning. (b) With consistency loss item in pre-training and fine-tuning. Dark blue and light blue denote time-based embedding $\bm{z}^{\textsc{t}}_i$ and frequency-based embedding $\bm{z}^{\textsc{f}}_i$, respectively.  Each circle denotes an `epilepsy' sample, i.e., the subject has a seizure episode, while each triangle denotes a sample associated with `health', i.e., the subject is healthy. The annotated $\bm{x}_i$ (marked as dashed line) is the same sample in two panels. 
    }
    \label{fig:visualization}
\end{figure}

\section{Visualization of embeddings in time-frequency space} 
\label{app:visualization}
To better demonstrate the effectiveness of the developed consistency loss $\mathcal{L}_{\textsc{c}}$, we visualize the learned embeddings in time-frequency space. In one-to-one setting, we pre-train the TF-C model on \textbf{\textsc{SleepEeg}} and fine-tune on \textbf{\textsc{Epilepsy}} (Sec.~\ref{sub:one_to_one}). We visualize the testing samples of \textbf{\textsc{Epilepsy}} in fine-tuning stage. The learned time-based embeddings $\bm{z}^{\textsc{t}}_i$ and  frequency-based embedding $\bm{z}^{\textsc{f}}_i$ have 128 dimensions. We map the embeddings to a 2-dimensional space through UMAP (20 neighbors; minimum distance as 0.2; cosine distance) for visualization. 

In this work, we develop a novel consistency loss $\mathcal{L}_{\textsc{c}}$ to induce the model learn closer time-based and frequency-based embeddings in a time-frequency space (Sec.~\ref{sub:TF-C}). As our problem setting is that the fine-tuning dataset has small scale ($M \ll N$; Sec.~\ref{sec:problem_formulation}), we randomly select 100 samples from the testing set (\textbf{\textsc{Epilepsy}}) for visualization.
The visualization of the learned embeddings in time-frequency space is shown in Figure~\ref{fig:visualization} (without $\mathcal{L}_{\textsc{c}}$ v.s. with $\mathcal{L}_{\textsc{c}}$). We can observe that: (1) when without $\mathcal{L}_{\textsc{c}}$ (panel a), the time-based and frequency-based embeddings are clustered separately; the embeddings are largely clustered together when having $\mathcal{L}_{\textsc{c}}$ (panel b). The observation demonstrates that the proposed $\mathcal{L}_{\textsc{c}}$ has the ability to push time-based and frequency-based embeddings closer to each other. (2) for a specific sample $\bm{x}_i$ (marked as dashed line), we can find that its time-based embedding (colored as light blue) and frequency-based (colored as dark blue) embedding are closer to each other in panel (b) than in panel (a). In particular, the cosine distance, between $\bm{z}^{\textsc{t}}_i$ and $\bm{z}^{\textsc{f}}_i$ of this sample, has reduced from 0.89 to 0.71 when using $\mathcal{L}_{\textsc{c}}$ in our model.
(3) the frequency-based embeddings of `epilepsy' and `health' samples are largely overlapped when without $\mathcal{L}_\textsc{c}$ but clustered well when considering $\mathcal{L}_\textsc{c}$, indicating the learned embeddings (with $\mathcal{L}_\textsc{c}$) can potentially enhance downstream classification performance.

\begin{figure}
    \centering
    \includegraphics[width=0.8\linewidth]{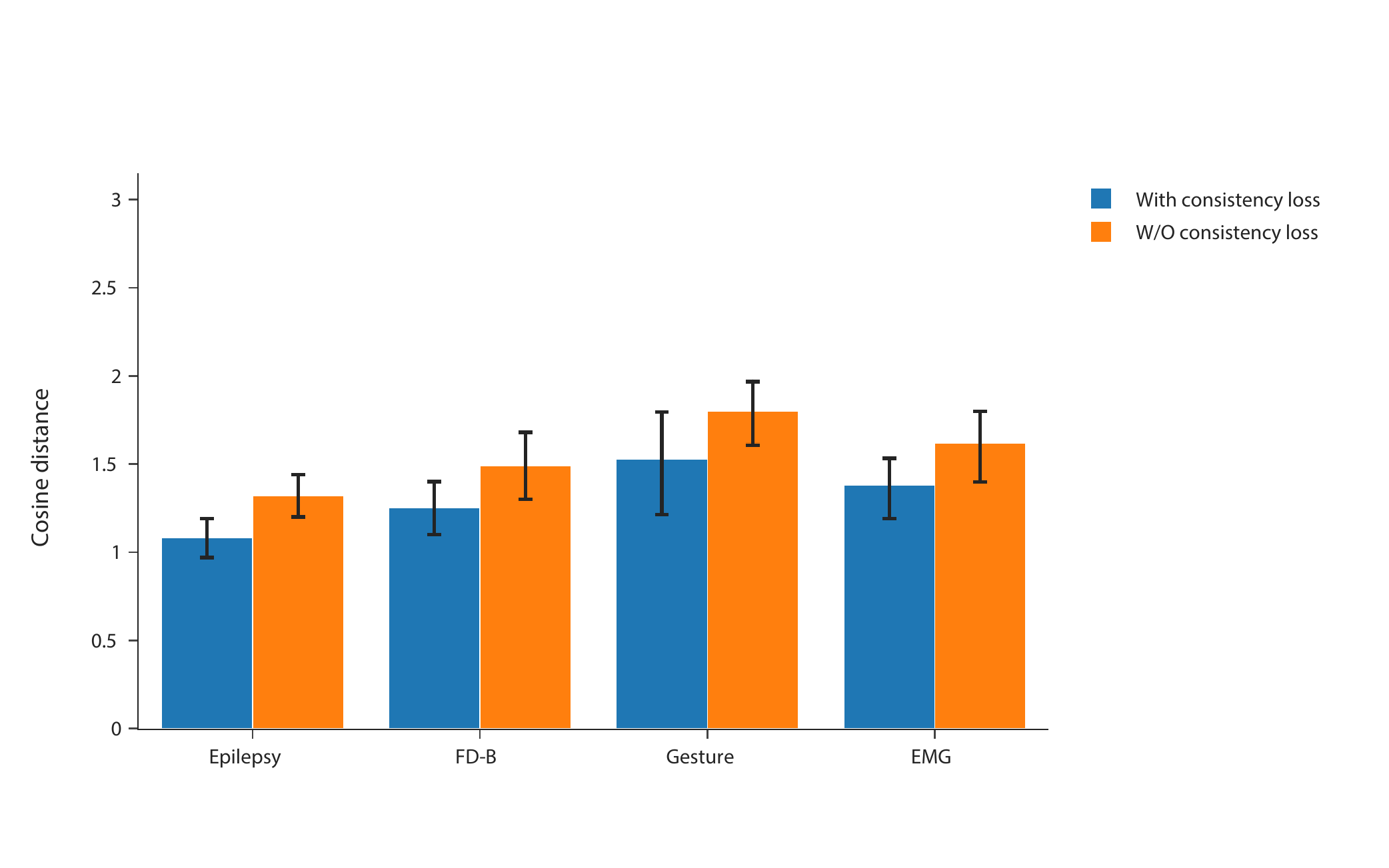}
    \caption{Cosine distance between \textit{time-based and frequency-based embeddings}. Here, lower distance is better as it denotes higher consistency between time-based and frequency-based embeddings. 
    The results are reported for the one-to-many setting, \ie, pretraining the model on \textbf{\textsc{SleepEeg}} dataset and fine-tuning on \textbf{\textsc{Epilepsy}}, \textbf{\textsc{Fd-B}}, \textbf{\textsc{Gesture}}, and \textbf{\textsc{EMG}}, respectively. In a specific dataset, \textit{for each sample}, we measure the cosine distance between its time-based embedding $\bm{z}^{\textsc{t}}_i$ and  frequency-based embedding $\bm{z}^{\textsc{f}}_i$ in a 128-dimension time-frequency space. In this figure, we report the cosine distance averaged across all testing samples. We observe that the cosine distance is larger when removing our proposed consistency loss item $\mathcal{L}_{\textsc{c}}$, indicating our model indeed pushes the time-based and frequency-based embeddings closer to each other.  
    }
    \label{fig:distance}
\end{figure}

\begin{figure}
    \centering
    \includegraphics[width=0.8\linewidth]{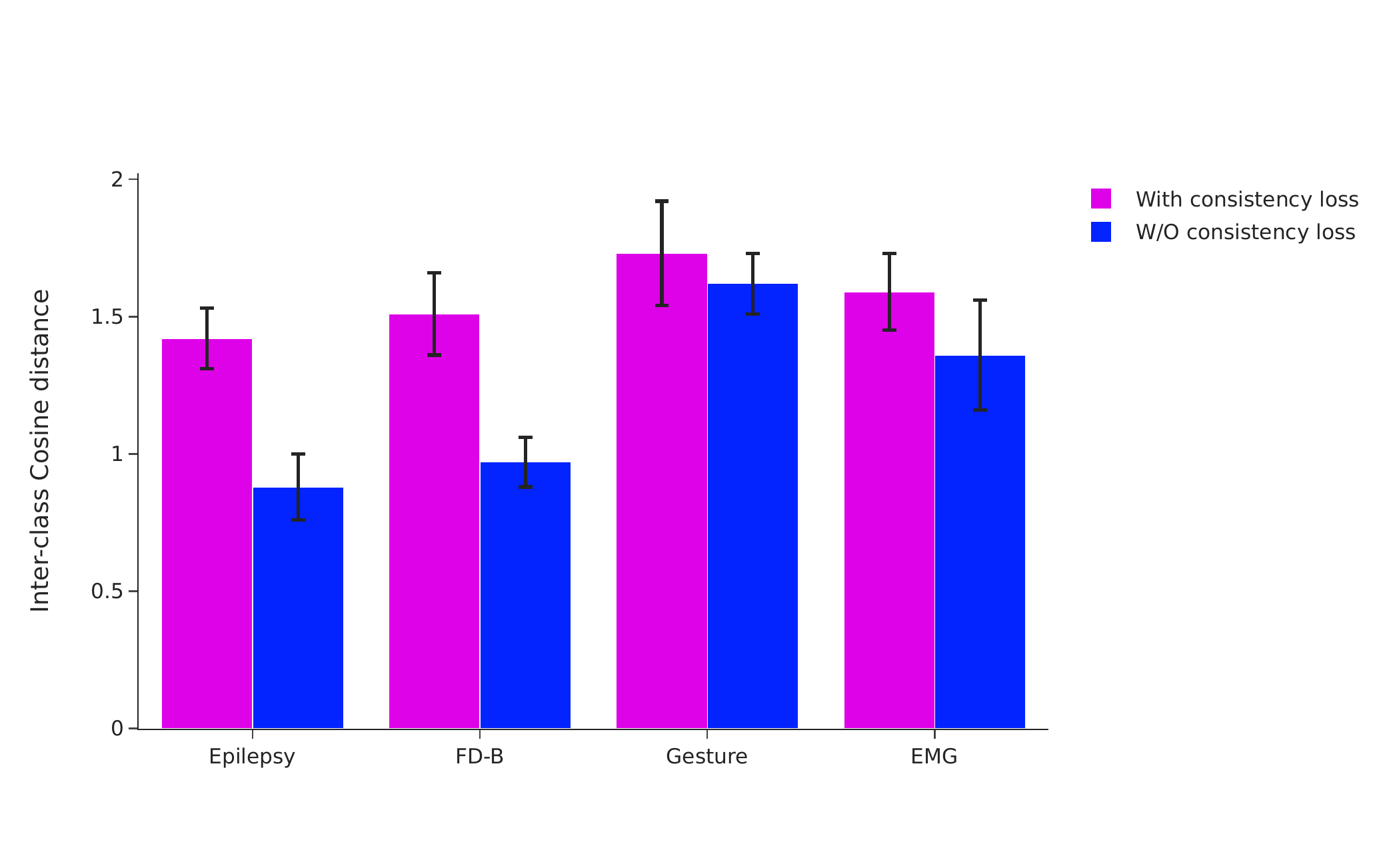}
    \caption{Cosine distance among \textit{embeddings of samples belonging to different classes}. The higher distance indicates better discrimination and better downstream classification performance. As in Figure~\ref{fig:distance}, the current experimental setup uses the one-to-many setting, but the calculation of cosine distance is slightly different. Take \textbf{\textsc{Epilepsy}} as an example, we first measure the distance between a positive sample and all the negative samples, then take the averaged distance across all positive samples as inter-class distance. The distance is calculated based on the concatenation of time-based and frequency-based embeddings, \ie, $\bm{z}_i^{\textrm{tune}}$. 
    We observe that our TF-C model (with consistency loss $\mathcal{L}_{\textsc{c}}$ increase the inter-class distance from 0.88 to 1.42 with a remarkable margin of 61.4\% (one-to-many setting; testing set of \textbf{\textsc{Epilepsy}}), indicating our model indeed has the ability to increase the distinction of learned embeddings and bring better classification performance on the fine-tuning dataset of interest. 
    The margins in \textbf{\textsc{Gesture}} and \textbf{\textsc{Emg}} are relatively low, and one potential reason is that the testing set of them are very small (120 for \textbf{\textsc{Gesture}}; 41 for \textbf{\textsc{Emg}}). 
    }
    \label{fig:interclass_distanc}
\end{figure}

Furthermore, we attempt to quantify the impact of our consistency loss item on the quality of learned embeddings, using the one-to-many setting as shown in Figure~\ref{fig:distance}. We pretrain the TF-C model on \textbf{\textsc{SleepEeg}} dataset and fine-tuning on \textbf{\textsc{Epilepsy}}, \textbf{\textsc{Fd-B}}, \textbf{\textsc{Gesture}}, and \textbf{\textsc{Emg}}, respectively. 
Next, we explain how the cosine distance is calculated taking \textbf{\textsc{Emg}} dataset as an example.
For every sample $\bm{x}_i$ in \textbf{\textsc{Emg}}, we measure the cosine distance between its time-based embedding $\bm{z}^{\textsc{t}}_i$ and  frequency-based embedding $\bm{z}^{\textsc{f}}_i$ in the 128-dimension time-frequency space. Then, we take the average of the distance across all testing samples and report in Figure~\ref{fig:distance}.
The number of testing samples of each dataset can be found in Table~\ref{tab:data_statistics}. We find that the cosine distance is larger when removing the proposed consistency loss item $\mathcal{L}_{\textsc{c}}$, which means our model, as designed and expected, indeed pushes the time-based and frequency-based embeddings closer to each other. For example, in \textbf{\textsc{Epilepsy}}, the averaged cosine distance dropped 22.2\% (from 1.32 to 1.08) when using consistency loss (Sec.~\ref{sub:TF-C}). At the same time, the embeddings of samples drawn from different classes are more easily distinguished from each other, as shown in Figure~\ref{fig:interclass_distanc}: we observe that our TF-C model (with consistency loss $\mathcal{L}_{\textsc{t}}$ increase the inter-class cosine distance from 0.88 to 1.42 with a remarkable margin of 61.4\% (one-to-many setting; testing set of \textbf{\textsc{Epilepsy}}), indicating our model indeed has the ability to increase the distinctiveness of learned embeddings and bring forward better classification performance on the fine-tuning dataset of interest. 
The results of visualization increase our confidence that the developed TF-C pre-training model can capture generalizable properties and boost knowledge transfer across different time series datasets.

\begin{figure}
    \centering
    \includegraphics[width=\linewidth]{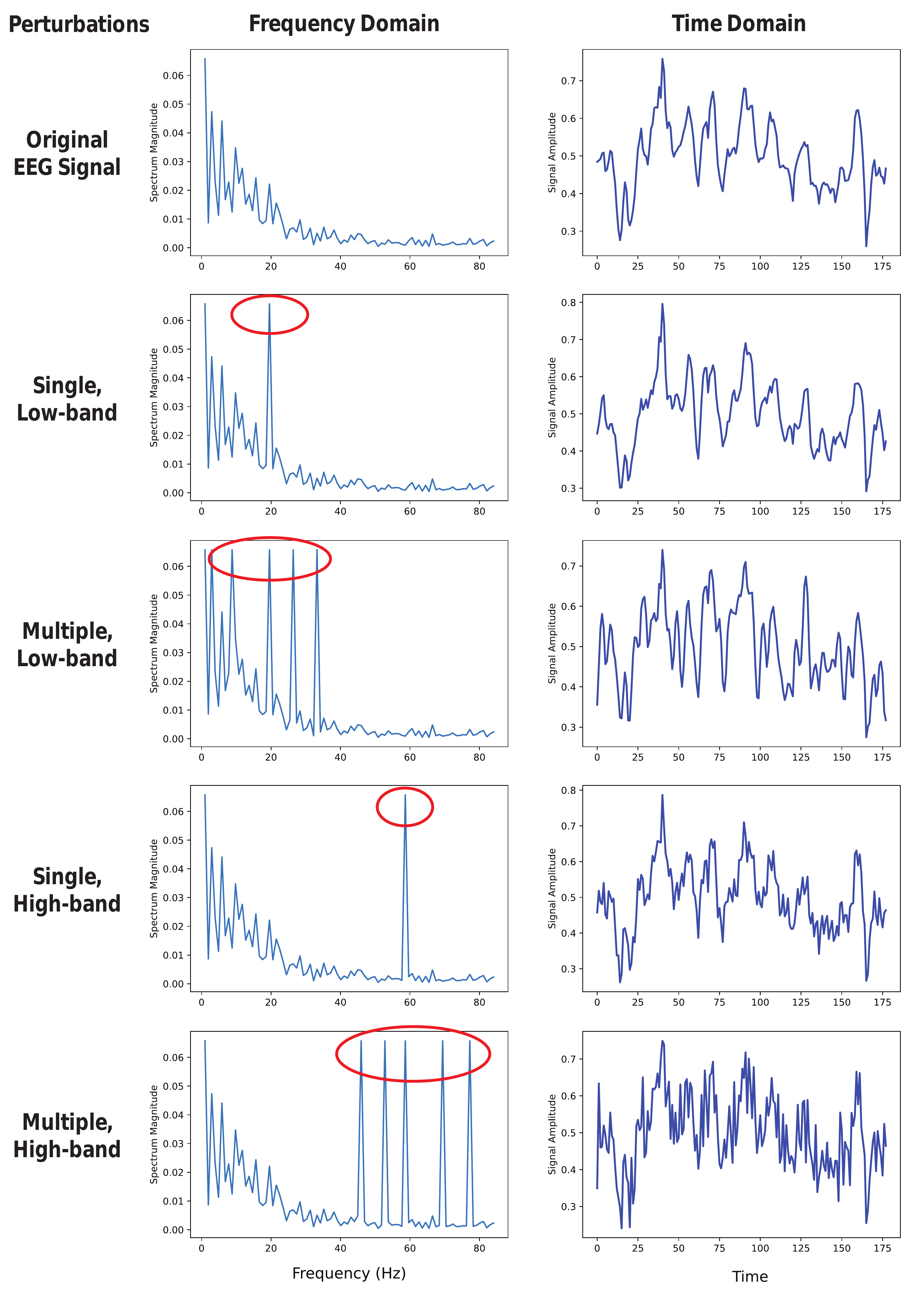}
    \caption{Illustration of augmentations in the frequency domain and the corresponding influence on the time domain. We considered the frequency augmentation from three aspects: low-band v.s. high-band, single-perturbation v.s. multiple perturbation, random selection v.s. distribution-based selection. The selection method has relative slight effect on time domain, thus, we mainly show how the perturbation budget (single or multiple) and frequency bands affect the temporal pattern in time domain. Here, for better visualization, we select $\alpha=1$ which means the perturbed component has the maximum magnitude (we set $\alpha=0.5$ in TF-C model) and set budget as 5 in multiple perturbation. The developed augmentations in this paper only perturb magnitude and keep phase information unchanged. We notice that perturbing phase is a crucial research direction in frequency domain augmentation in the future work.
    }
    \label{fig:freq_aug}
\end{figure}

\section{Additional information on domain augmentations} 
\label{app:domain_augmentation}

To the best of our knowledge, this is the first work that direct perturbs the frequency domain of time series in contrastive learning. Here, we thoroughly explore various frequency augmentations and discuss the design of manipulation strategy in frequency domain. In particular, we analyze eight frequency augmentation policies and discuss their performance. 

We consider three types of frequency augmentations: 
\begin{itemize}
    \item \textbf{Low- vs. High-band perturbations}: High-frequency components contribute to the fast varying parts of the time series (e.g., sharp transitions). In contrast, low-frequency components contribute to slow signal variations in the time domain. Low band perturbations correspond to the first half of the frequency spectrum. 
    \item \textbf{Single- vs. Multi-component perturbations} (aka varying budget size): Perturbing several frequency components (i.e., larger budget) leads to more substantial changes in the time domain. In this analysis, we set the budget to 5 for multi-component perturbations and 1 for single-component perturbations. 
    \item \textbf{Random vs. Distributional perturbations}: The selection of components in the spectrum that get perturbed is important because it influences the resulting time-based representation of the augmented sample. Frequency components can be sampled uniformly at random or according to a user-specified distribution (i.e., Gaussian in the following analysis). 
\end{itemize}

Based on the above considerations, we developed eight frequency augmentations. 
In Fig.~\ref{fig:freq_aug} (\textsc{Epilepsy} dataset)), we visualize a series of perturbations in the frequency domain and their effects in the time domain (through the inverse FFT).

In addition, we explore how different spectral perturbations affect the performance of TF-C. Table~\ref{tab:spectral} shows results for the one-to-one setup (\textsc{SleepEeg} $\rightarrow$ \textsc{Epilepsy}). We find the following:

\begin{itemize}
    \item Perturbing low-band components performs better than augmenting high-band elements of the spectrum (0.8\% change in F1 performance on average across the four setups). This can be explained by the nature of the underlying time series: in physiological signals, most information is carried in a low band (e.g., EEG signals are most informative in 0.5-70 Hz while > 70Hz is considered noise). This means that the choice of low- vs. high-band components depends on the signal context. For example, high-band components are informative for mechanical signals. We further verified this by external experiments in the \textsc{Fd-a} $\rightarrow$ \textsc{Fd-b} setting, where perturbing high-band components outperform low-band augmentations by 0.8\% in F1. Thus, we consider perturbations across the entire spectral band in the paper to achieve broad generalization.
    \item In general, perturbing many frequency components degrades model performance. One reason for worse performance is that too big a budget incurs too significant changes in the time-based representation of augmented samples. In that case, the augmented samples would be easily distinguished by a contrastive model, leading to poor contrastive encoders. For this reason, we use single-component perturbations in the paper.
    \item We cannot draw a clear conclusion on whether the random selection of components is better than Gaussian distribution-based as both F1 scores are very similar. The optimal choice of sampling strategy likely depends on both the time-series signal and prediction task. We use random sampling in the paper.
\end{itemize}

\begin{table}[]
\centering
\scriptsize
\def\arraystretch{1.0}
\caption{\textbf{Results comparison of eight frequency augmentations.} Considering the perturbing bands, the number of components, and the perturbation distribution, there are eight different combinations. Here we regard each factor as binary (e.g., Low v.s. High), we note that the more fine-grid factors (e.g., Delta band v.s. Alpha band v.s. Beta band v.s. Gamma band in the context of EEG signals) might be helpful for augmentation design.
}
\label{tab:spectral}
\begin{tabular}{lllll}
\toprule
\textbf{No.} & \textbf{Low vs. High} & \textbf{Single vs. Multi} & \textbf{Random vs. Gaussian} & \textbf{F1 score} \\ \midrule
1 & Low & Single & Random & 0.9136 \\
2 & Low & Single & Gaussian & 0.8957 \\
3 & Low & Multi & Random & 0.8763 \\
4 & Low & Multi & Gaussian & 0.8859 \\
5 & High & Single & Random & 0.8972 \\
6 & High & Single & Gaussian & 0.8651 \\
7 & High & Multi & Random & 0.8818 \\
8 & High & Multi & Gaussian & 0.8775\\   
\bottomrule
\end{tabular}
\end{table}

\section{Discussion on many-to-one setting}
\label{app:many-to-one}
We probed the one-to-one setting and one-to-many setting in knowledge transfer of time series (Sec.~\ref{sec:experiments}). Here, we discuss the many-to-one set up: pre-train on a mixture of multiple datasets and fine-tune on a single pure dataset.

We notice that the many-to-one setup fundamentally differs from the one-to-one and also from one-to-many configurations. 
Notably, in the one-to-one and one-to-many setups, the model is pre-trained on a single pre-training dataset presumed to originate from a data source exhibiting homogeneity. For example, samples in the pre-training dataset could describe similar devices (e.g., different kinds of boring machines) with similar sampling rates, lengths, semantic meanings, and/or relevant temporal patterns that occur on comparable time scales, etc. When some level of homogeneity exists in a pre-training dataset, the method can learn patterns and apply them to different fine-tuning datasets. 
In contrast, in many-to-one setups, pre-training datasets can be incredibly heterogeneous. For example, a pre-training dataset might contain samples describing boring machines' behavior and recordings of EEG medical devices. For this reason, pre-training can be severely hampered by diversity and misalignments across samples in the pre-training dataset. When we merge many heterogeneous samples into a single pre-training dataset, the model might struggle to find patterns underlying the combined dataset. While models in computer vision have been successfully pre-trained on large, diverse datasets, such as ImageNet, the literature on pre-training for time series is much sparser. Further, heterogeneous pre-training datasets pose additional challenges due to the unique properties of time series. Learning universal representations from multi-sourced time series datasets (i.e., many-to-one setups) remains an open challenge. It is an interesting future direction, and we hope this study can facilitate future research. 

We have explored this and provide preliminary experimental results in four many-to-one setups (named N-to-one for N = 1,2, 3, and 4). 
Take four-to-one as an example. We merge four pre-training datasets (\textsc{SleepEeg},\textsc{Fd-a}, \textsc{Har}, and \textsc{Ecg}) into one pre-training dataset. We had to preprocess the datasets to make a many-to-one setup feasible. To address the mismatch between the number of channels, for the multivariate dataset (\textsc{Har}), we take a single channel (i.e., x-axis acceleration) to make sure all samples have the same number of channels. To align the length across datasets, we make all samples have 1,500 observations by zero-padding (for the ones shorter than 1,500) and cutting-off (for the ones longer than 1,500). Further, we take 25\% samples (instead of 100\%; for computational reasons only) from each of the four datasets to create a pre-training dataset with 108,300 samples that belong to 18 classes in total. Similarly, we built pre-training for three-to-one (merge \textsc{SleepEeg},\textsc{Fd-a}, and \textsc{Har}) and two-to-one (merge \textsc{SleepEeg} and \textsc{Fd-a}). 

We take the above N-to-one pre-training datasets and repeat the following analysis. First, we pre-train our TF-C and then fine-tune it on \textsc{Epilepsy} for classification. The AUROC for one-to-one, two-to-one, three-to-one, and four-to-one settings are 0.9819, 0.9525, 0.8607, and 0.7253, respectively.
We observe AUROC decreases with the increasing heterogeneity of pre-training datasets, which is consistent with our analysis in the first part of this answer. Nevertheless, whether there exists a framework that overcomes the problems and enables many-to-one pre-training for time series remains to be seen, and we believe it's an exciting direction for future research.

\clearpage
\bibliographystyleS{unsrt} 
\bibliographyS{SIrefs}
\end{appendices}



\end{document}